\documentclass[lettersize,journal]{IEEEtran}
\usepackage{amsmath,amsfonts}
\usepackage{algorithmic}
\usepackage{array}
\usepackage{textcomp}
\usepackage{stfloats}
\usepackage{url}
\usepackage{verbatim}
\usepackage{graphicx}
\def\BibTeX{{\rm B\kern-.05em{\sc i\kern-.025em b}\kern-.08em
    T\kern-.1667em\lower.7ex\hbox{E}\kern-.125emX}}
\usepackage{balance}

\usepackage{amsmath,amssymb,amsfonts,amsthm}
\usepackage{algorithmic}
\usepackage{graphicx,color}
\usepackage{textcomp}
\usepackage{xcolor}
\usepackage{hyperref}
\hypersetup{hidelinks=true}
\usepackage{algorithm,algorithmic}
\usepackage{siunitx}

\AtBeginDocument{\definecolor{tmlcncolor}{cmyk}{0.93,0.59,0.15,0.02}\definecolor{NavyBlue}{RGB}{0,86,125}}

\usepackage[style=ieee,sorting=none]{biblatex}
\usepackage{tikz}
\graphicspath{{img/}}
\def\R{\mathbb{R}}
\def\N{\mathbb{N}}

\newcommand{\Norm}[1]{\left\lVert #1 \right\rVert}

\newcommand{\ceil}[1]{\lceil #1 \rceil}
\newcommand{\floor}[1]{\lfloor #1 \rfloor}

\newcommand{\Cauchy}{\operatorname{Cauchy}}
\newcommand{\Bernoulli}{\operatorname{Bernoulli}}
\DeclareMathOperator*{\argmin}{arg\,min}

\DeclareMathOperator{\sign}{sign}

\DeclareMathOperator{\Laplace}{Laplace}
\DeclareMathOperator{\Median}{Median}

\DeclareMathOperator{\Uniform}{U}

\newcommand*\diff{\mathop{}\!\mathrm{d}}

\def\Proj{\operatorname{Proj}}
\def\L1{\ensuremath{L_1}}
\def\Lp#1{\ensuremath{L_{#1}}}

\def\MedReg{M}
\def\LoMed{L_\text{low}}
\def\HiMed{L_\text{hi}}
\def\kLev#1#2{\alpha_{#1}({#2})}
\def\kLevBound#1#2{\hat{\alpha}_{#1}({#2})}

\newtheorem{thm}{Theorem}[section]
\newtheorem{lemma}[thm]{Lemma}
\newtheorem{prop}[thm]{Proposition}

\definecolor{myieeeblue}{HTML}{226FA9}
\usepackage{siunitx}
\usepackage{xspace}

\usepackage[font=footnotesize,labelfont=bf]{caption}
\usepackage[font=footnotesize,labelfont=bf]{subcaption}
\usepackage{microtype}

\begin{document}

\title{Fast and Exact Least Absolute Deviations Line Fitting \\ 
via Piecewise Affine Lower-Bounding}

\author{Stefan Volz, Martin Storath, Andreas Weinmann
\thanks{The authors are with the Institute of Digital Engineering and the Faculty of Applied Natural Sciences and Humanities, Technische Hochschule Würzburg-Schweinfurt, Schweinfurt, Germany
}
\thanks{AW acknowledges support by Deutsche Forschungsgemeinschaft (DFG, German Research Foundation) INST 168/3-1, INST 168/4-1.
Further, his work was supported by the Project AI4WildLIVE funded by the German Federal Ministry for Research, Technology and Space (BMFTR) under 03LW0652. The responsibility for the contents of this publication lies with the authors.

ChatGPT (OpenAI, last accessed Dec. 12, 2025) was used to refine English grammar and suggest alternative phrasings. The authors checked all suggestions for technical accuracy and take full responsibility for the final text.
}
}

\maketitle

\begin{abstract}

Least-absolute-deviations (LAD) line fitting is robust to outliers but computationally more involved than least squares regression. Although the literature includes linear and near-linear time algorithms for the LAD line fitting problem, these methods are difficult to implement and, to our knowledge, lack maintained public implementations. As a result, practitioners often resort to linear programming (LP) based methods such as the simplex-based Barrodale-Roberts method and interior-point methods, or on iteratively reweighted least squares (IRLS) approximation which does not guarantee exact solutions. 
To close this gap, we propose the Piecewise Affine Lower-Bounding (PALB) method, an exact algorithm for LAD line fitting. PALB uses supporting lines derived from subgradients to build piecewise‑affine lower bounds, and employs a subdivision scheme involving minima of these lower bounds.
We prove correctness and provide bounds on the number of iterations. On synthetic datasets with varied signal types and noise including heavy-tailed outliers as well as a real dataset from the NOAA's \emph{Integrated Surface Database},
PALB exhibits empirical log-linear scaling. 
It is consistently faster than publicly available implementations of LP based and IRLS based solvers. We provide a reference implementation written in Rust with a Python API.

\end{abstract}

\begin{IEEEkeywords}
  Algorithms, Computational efficiency, Convex functions, Iterative algorithms, Linear regression, Mathematical programming, Noise robustness, Optimization
\end{IEEEkeywords}

\maketitle

\begin{refsection}
  \section{Introduction}

\IEEEPARstart{F}{itting} a line to a signal or a set of data points is a fundamental task in natural sciences and engineering.
Examples are signal denoising,  trend estimation, or sensor data regression, and it is most frequently performed via least squares estimation. In the presence of outliers, least absolute deviations (LAD) fitting provides a more robust alternative.
Given data
${\{(x_i, y_i)\}}_{i=1,...,N} \subset \R^2$
the LAD line fitting problem is formulated as follows:
\begin{align}\label{eq:main-problem}
  \argmin_{m,t \in \R} f(m,t), \quad \text{with } f(m,t) = \sum_{i=1}^N |mx_i + t - y_i|. \tag{$\star$}
\end{align}
Unlike the least squares solution, the LAD estimator lacks a closed-form expression,
and iterative solvers have to be applied.

The solution of \eqref{eq:main-problem} is not necessarily unique, but an optimal line always passes through at least two points. In principle, the problem could be solved by enumerating all point pairs and computing the corresponding objective value which results in cubic time complexity. A more structured alternative reformulates it as a linear program (LP) with primal or dual forms~\cite{armstrong1982dual}. Barrodale and Roberts~\cite{barrodale1973improved, BarrodaleRoberts_1974} developed a specialized simplex-type algorithm with quadratic time complexity, implemented in libraries such as L1pack~\cite{osorio2025routines} and quantreg~\cite{koenker2009quantreg}. Megiddo and Tamir~\cite{megiddo1983finding} proposed a near-linear complexity algorithm using subgradient information and a specialized sorting scheme, and Yamamoto et al.~\cite{yamamoto1988algorithms} achieved linear complexity through iterative pruning. Although theoretically optimal or near-optimal, these latter algorithms are challenging to implement and have seen limited practical adoption, and LP-based solvers remain the dominant exact methods in practice. Approximate approaches such as iteratively reweighted least squares (IRLS)~\cite{schlossmacher1973iterative} are simple to implement but do not guarantee exact solutions.

Publicly available implementations mainly rely on LP formulations,
which are solved via simplex or interior-point methods, or on IRLS for approximate solutions. Packages such as quantreg, L1pack, scikit-learn, and statsmodels cover both families, with LP-based solvers offering exactness but limited scalability and IRLS providing faster convergence at the cost of accuracy. The relative performance depends strongly on data size and structure~\cite{Gentle_1988}, which becomes critical when solving many LAD problems of varying lengths, for instance in large-scale regression tasks or in time-series segmentation~\cite{killick2012optimal,weinmann2014l1potts,DofPPR}.

Despite a large body of research, the LAD line fitting problem still lacks a solver that is simultaneously fast, exact, and implementable with standard numerical schemes, with an implementation 
that is efficient across diverse dataset sizes. 

\subsection{Prior and Related Work}
\label{sec:stateoftheart}

LAD line fitting can be tracked back to the 18th century, even before least squares fitting.
A brief historical account can be found in \cite{Li_2004}.

Algorithmic approaches can be grouped as in the following list. Note that some methods can be applied to more than two variable problems, so we distinguish between the two-parameter case (slope and intercept) \emph{LAD line fitting}  and the general case with multiple predictors, which we refer to as \emph{LAD linear regression.}

\subsubsection{Linear-programming methods}
As already mentioned, the problem \eqref{eq:main-problem} admits reformulation in primal and dual LP forms which can be solved with generic LP solvers. Both simplex or  interior-point algorithms are used in practice.
As already mentioned,
Barrodale and Roberts~\cite{barrodale1973improved, BarrodaleRoberts_1974} have proposed a simplex-type algorithm for LAD linear regression; Armstrong and Kung \cite{armstrong1978algorithm} have devised a specialized version of this algorithm for the LAD line fitting problem.
The Frisch-Newton interior point method, implemented in the quantreg package, has been reported to be advantageous over simplex type methods for large data sets \cite{portnoy1997gaussian}. 

\subsubsection{Weighted Median Methods} 
These type of methods exploit the fact that 
weighted medians can be computed in linear time \cite{bleich1983linear,rauh2012optimal}.
Karst~\cite{karst1958linear}   computes the exact LAD line constrained to pass through a designated point by a single weighted-median step, and iteratively updates the designated point.
 A similar computing scheme is used in \cite{Sabo_2008}.
 Bloomfield and Steiger \cite{Bloomfield_1980}
propose to choose a coordinate of steepest normalized descent and solve
the resulting one-dimensional subproblem exactly by a weighted median.
Wesolowsky's~\cite{Wesolowsky_1981} method
is based on the observation that a minimum can be found at the intersection of two zero residual lines $mx_i + t - y_i = 0,$ $i=1,\ldots, N,$  
in $m, t$-space. The algorithm moves along the steepest-descent lines and solves the associated one-dimensional subproblem by a weighted median, where a special handling of multiple intersecting lines is needed.
 Josvanger and Sposito~\cite{josvanger1983l1norm} proposed a variant of  Wesolowsky's method, and reported lower runtimes than the simplex-based implementation of~\cite{armstrong1978algorithm}.
 Li and Arce~\cite{Li_2004} 
refined that method by a coordinate transformation to direct the optimization procedure along edge lines of the cost surface.

\subsubsection{Specialized Exact Methods for LAD Line Fitting}
Megiddo and Tamir \cite{megiddo1983finding}
proposed using subdifferential information 
for $g(m) = \min_{t \in \R} f(m,t)$ and a parallel sorting scheme \cite{preparata1978new}
applied sequentially to be able to sort by an unknown key, which is the optimal slope. Its time complexity is linearithmic in the number of data points.
The method of Yamamoto et al.~\cite{yamamoto1988algorithms}
uses an iterative pruning method. It builds   on the work of Meggido \cite{megiddo1984linear} for linear programming.

\subsubsection{Approximate Methods}
Schlossmacher~\cite{schlossmacher1973iterative}
proposed iteratively reweighted least squares (IRLS) for LAD linear regression.
A variant of the iteratively reweighted least squares using a Gauss-Newton iteration has been proposed in \cite{Sabo_2008}.
While simple and often fast, IRLS does not in general guarantee exact  solutions.
For small scale data, LP solvers might be even faster, cf. \cite{armstrong1976comparison} and the experimental section of this work.

\subsubsection{Software Implementations}
Although the algorithms of Megiddo-Tamir and Yamamoto have theoretically optimal or near-optimal scaling they are challenging to implement, and the authors are not aware of maintained publicly available implementations of these methods.
Instead, the R packages L1pack~\cite{osorio2025routines}
and quantreg~\cite{koenker2009quantreg} rely on the
the  Barrodale-Roberts method, and quantreg additionally implements an   interior point LP solver \cite{koenker2005frisch} for data with more than 5\,000 points. (Note that LAD fitting is equivalent to 50\% quantile regression.)
The Python library scikit-learn uses the LP library HiGHS to solve an LP formulation.
The Python library statsmodels uses  iteratively reweighted least squares \cite{schlossmacher1973iterative}.

\subsection{Contribution}
We propose a new algorithm, termed the Piecewise Affine Lower-Bounding (PALB) method, for solving the LAD problem~\eqref{eq:main-problem} exactly.
Following the reformulation of Megiddo and Tamir~\cite{megiddo1983finding}, we consider the convex, piecewise-linear function
$J(m) = \min_{t \in \R} f(m,t),$
and exploit the fact that its subdifferential can be computed in linear time.

The key novelty of our method is using this subdifferential to construct a piecewise affine lower bound
over a given interval of slopes $[m_l, m_r].$
This lower bound is formed by the two supporting lines at the boundaries $m_l$ and $m_r$  that are the largest possible lower bounds to the objective function on $[m_l, m_r]$ that consists of two linear pieces.
The minimizer of this  function $m_s$ (the intersecting point of the two lines)
subdivides the interval, and the sign of the subgradient at this point indicates on which side of $m_s$ the sought minimizer $m^*$ lies.
This procedure is repeated on the subdivided interval until zero is in the subdifferential,
which gives the sought minimum $m^*.$
(Once $m^*$ is found, the optimal intercept $t^*$ is obtained by computing a median of $\{y_i - m^* x_i\}_{i=1}^N.$)

This procedure can be viewed as a type of bisection method using subgradient information.
However, if we used a fixed rule for choosing the cut point, such as midpoint or golden ratio,
termination with a minimizer would not be guaranteed, and therefore would only lead to an approximate solver.
The crucial point of PALB is that is chooses the cut point as the minimizer
of the current piecewise affine lower bound (the intersection of the endpoint supports),
which assures finite termination at a minimizer.

We give proofs of correctness for this method and give bounds on its complexity;
in particular we show that PALB yields global optimizers in a finite number of steps.
Moreover, we derive both concrete and asymptotic bounds on the worst-case and expected number of iterations.
These bounds are obtained by relating the LAD problem to the classical $k$-level problem in combinatorial geometry~\cite{Erdos_1973, Lovasz_1971}.
Empirical results indicate a log-linear time complexity. 

As the methods of \cite{megiddo1983finding} and Yamamoto et al. \cite{yamamoto1988algorithms}, PALB leverages the subdifferential of $g,$ but the search procedure is much simpler as it essentially consists of computing the intersection of two straight lines.
In contrast, the Megiddo–Tamir method relies on Preparata’s parallel sorting scheme, and Yamamoto et al. require intricate pruning and bookkeeping of a feasible set.

The PALB core is implemented in Rust, with a user‑friendly Python interface for ease of use. The current implementation processes more than a million samples in under one second on a standard workstation and features a static memory profile, making it suitable for embedded and real‑time applications.
The source code for our implementation is hosted on GitHub at~\url{https://github.com/SV-97/piecewise-affine-lower-bounding}.

We benchmark our method against several state‑of‑the‑art solvers for \eqref{eq:main-problem}, focusing on methods with publicly available implementations including the Barrodale–Roberts algorithm (L1pack), LP formulations solved by simplex methods (IBM CPLEX), an interior‑point method (Clarabel), and the approximate IRLS method (statsmodels).
(Due to the lack of publicly available implementations, runtimes for the methods of Megiddo-Tamir and Yamamoto et al. could not be reported.)
The results demonstrate that PALB achieves substantially lower median runtimes than all competing methods on simulated data ranging from ten to several million samples.
Runtime performance profiles further show that PALB is the fastest exact method in nearly all tested real-world scenarios.

\subsection{Organization of the Paper}
In Section~\ref{sec:palb} we describe the proposed Piecewise Affine Lower-Bounding Method (PALB) method
and analyze its complexity.
In Section~\ref{sec:impl-and-experiments} we discuss
the implementation of our method and study several numerical experiments.

  \section{Piecewise Affine Lower Bounding Algorithm}\label{sec:palb}

\subsection{Method description}

The input data of the proposed method 
is a set of $N$ data points ${\{(x_i, y_i)\}}_{i=1,...,N} \subset \R^2,$
which are not required to satisfy any special assumptions, such as sortedness or uniqueness of the $x$-values.

The proposed method solves the LAD line fitting problem~\eqref{eq:main-problem} by minimizing the marginal function $J(m) = \min_{t \in \R} f(m,t)$. 
It is well known that the inner parametric problem is solved by finding a median $t^*(m) \in \Median{\{y_i - m x_i\}}_i$ of the residuals ${\{y_i - m x_i\}}_i$, so that it actually becomes practically feasible to reduce the minimization of $f$ to that of $J$.
The function $f$ is convex, continuous, coercive and piecewise affine, and $J$ inherits all of these properties so that in particular its subdifferential $\partial J$ is a monotone mapping.
Notably this subdifferential $\partial J$ can actually be fully computed in practice --- we describe a linear time method for doing so in supplementary Section~\ref{subsec:subdiff-J}. It is principally a problem that can be solved by standard algorithms that are often times used to compute medians, such as the \emph{median of medians} of \cite{BlumFloydPrattRivestTarjan1973} or introselection methods \cite{Musser1997, Alexandrescu16} (in the following we refer to such linear time introselection methods simply as \emph{introselect}).
We leverage this specific structure of $J$ to implement what is essentially a highly specialized cutting-plane method adapted to the \L1 problem.
The piecewise linear structure of $J$ and an appropriate selection of subgradients then guarantees termination in finitely many steps and it is in fact possible to give upper bounds on the number of iterations as we show later.

An ordinary cutting-plane method (see for example \cite[Chapter XV, Section 1]{HiriartUrruty1993}) solves $\min_{x \in C} J(x)$ for a compact convex set $C$ and convex, continuous function $J$ by starting from a single initial guess $\xi_0 \in C$ and corresponding subgradient $g_0 \in \partial J(\xi_0)$. One then minimizes the affine map $J^{(0)} : x \mapsto J(\xi_0) + g_0(x - \xi_0)$ over $C$ to obtain a new guess $\xi_1$. Using this one chooses a subgradient $g_1 \in \partial J(\xi_1)$, minimizes $J^{(1)} : x \mapsto \max\{ J^{(1)}(x), J(\xi_1) + g_1(x - \xi_1) \}$ over $C$ to obtain $\xi_2$ and so on.

The proposed method differs from this basic scheme in that we begin by determining a suitable set $C = [a_0, b_0]$, and then maintaining a bracketing interval $[a_k, b_k]$ throughout, rather than a single trajectory of estimates.
We first identify two initial points $a_0 < b_0$ with opposing subgradient signs (i.e.\  $\partial J(a_0) < 0 < \partial J(b_0)$, which is to be understood as an elementwise inequality). This guarantees the interval $[a_0, b_0]$ contains all stationary points of $J$.
Subsequently, for each iteration $k \in \N_0$, we minimize the piecewise affine global minorizer 
\begin{align*}
  J^{(k)}(m) = \max_{\substack{i=0,...,k \\ \xi \in \{a_i, b_i\}}} J(\xi) + g_{\xi}(m - \xi)
  \\ \text{where}\quad g_{a_i} = \max \partial J(a_i), \quad g_{b_i} = \min \partial J(b_i)
\end{align*}
specifically over the current interval $[a_k, b_k]$.
This restriction simplifies the subproblem significantly, because on $[a_k, b_k]$ we have $J^{(k)}(m) = \max\{ J(a_i) + g_{a_i}(m - a_i), J(b_i) + g_{b_i}(m - b_i) \},$
such that the unique minimizer $\xi_k$ of $J^{(k)}$ is that point $\xi$, where
$J(a_k) + g_{a_k} (\xi - a_k) = J(b_k) + g_{b_k} (\xi - b_k)$.
Depending on the subdifferential $\partial J(\xi_k)$ we either terminate or update the intervals by replacing one of the two boundaries by $\xi_k$ such that the condition $\partial J(a_{k+1}) < 0 < \partial J(b_{k+1})$ is maintained for the next step.
Crucially, the piecewise linear structure of $J$ ensures that the minimizer of the approximation $J^{(k)}$ eventually coincides with that of the true objective, guaranteeing exact termination in finitely many steps.

For a robust implementation this core algorithm can be augmented with a safeguarding step:
while impossible in theory, in practice it could happen that, due to numerical inaccuracies, the new boundary point $\xi_k = \argmin_{x \in [a_k,b_k]} J^{(k)}(x)$ coincides with either $a_k$ or $b_k$. In this case the algorithm would get stuck.
To prevent this we use the following safeguarding mechanism. We determine a collection of intervals $\{S_k\}_k$ with $S_k \subseteq [a_k, b_k]$ and use the projection $\Proj_{S_k}(\argmin_{x \in [a_k,b_k]} J^{(k)}(x))$ instead of $\argmin_{x \in [a_k,b_k]} J^{(k)}(x)$ to determine the next interval $[a_{k+1}, b_{k+1}]$.
An appropriate choice of the $S_k$, cf.\ section~\ref{sec:impl-and-experiments}, further guarantees that the intervals $[a_k,b_k]$ shrink at some minimal rate.
By setting $S_k = [a_k,b_k]$ we get back the core method.

We note that, in principle, the proposed algorithm, cf.\ Algorithm~\ref{alg:main},
is applicable to any single-variable,
piecewise affine, continuous, convex function for which subdifferentials and values can be computed.

\begin{algorithm}[h]
  \caption{
    Safeguarded PALB algorithm.
    The sequence $\delta_i$ controls how exactly the interval $[a_k,b_k]$ is expanded,
    while the intervals $\{S_i\}$ realize the safeguard.
  }\label{alg:main}
  \begin{algorithmic}
    \STATE
    \STATE {\textsc{Safe-PALB}}$(a_k,b_k, {\{\delta_i\}}_{i=1,...}, {\{S_i\}}_{i=1,...})$
    \STATE \hspace{0.5cm}$ \textbf{compute } \partial J(a_k),~\partial J(b_k)$
    \STATE \hspace{0.5cm}$ \textbf{if either contains } 0$
    \STATE \hspace{1.0cm}$ \textbf{return } \text{stationary point}$
    \STATE \hspace{0.5cm}$ \textbf{else if both have opposing uniform signs}$
    \STATE \hspace{1.0cm} {\color{gray} Subdivide interval $[a_k,b_k]$}
    \STATE \hspace{1.0cm}$ \textbf{determine approximation } J^{(k)} \textbf{ to } J$
    \STATE \hspace{1.0cm}$ \xi_k \gets \Proj_{S_k}(\argmin_{x \in [a_k,b_k]} J^{(k)}(x))$
    \STATE \hspace{1.0cm}$ \textbf{if } \sign(\partial J(\xi_k)) = \sign(\partial J(a_k))$
    \STATE \hspace{1.5cm}$ a_{k+1} \gets \xi_k,~b_{k+1} \gets b_k$
    \STATE \hspace{1.0cm}$ \textbf{else if } \sign(\partial J(\xi_k)) = \sign(\partial J(b_k))$
    \STATE \hspace{1.5cm}$ b_{k+1} \gets \xi_k,~a_{k+1} \gets a_k$
    \STATE \hspace{1.0cm}$ \textbf{else it contains } 0$
    \STATE \hspace{1.5cm}$ \textbf{return } \xi_k$
    \STATE \hspace{0.5cm}$ \textbf{else both have same uniform sign}$
    \STATE \hspace{1.0cm} {\color{gray} Expand and shift interval toward minimizer}
    \STATE \hspace{1.0cm}$ d_k \gets -\sign(\partial J(a_k))\ {\color{gray} = -\sign(\partial J(b_k))}$
    \STATE \hspace{1.0cm}$ (a_{k+1}, b_{k+1}) \gets
    \begin{cases}
      (b_k, b_k + d_k \delta_k) & d_k = 1 \\
      (a_k + d_k \delta_k, a_k) & d_k = -1
    \end{cases}$

    \STATE \hspace{0.5cm}$ \textbf{return }{\textsc{Safe-PALB}}(a_{k+1}, b_{k+1}, \{\delta_i\}_{i=1,...}, {\{S_i\}}_{i=1,...})$
  \end{algorithmic}
\end{algorithm}

We speak of the \emph{uniform} sign of $\partial J(x)$ to emphasize that despite $\partial J(x)$ being a set, there is one unambiguous sign that can be assigned to the whole set at this point: either all its elements are positive, or all of them are negative.
We call steps taking the second branch in Algorithm~\ref{alg:main} \emph{subdivision steps},
and ones taking the third branch \emph{expansion steps}.
The expansion steps essentially perform a subgradient-directed search --- or as an alternative interpretation:
they take subgradient steps with both interval boundaries
in such a way that $[a_k,b_k] \cap [a_{k+1},b_{k+1}]$ is always a singleton.

The choice of the values $\delta_k$ and starting points $a_0, b_0$ impacts how fast the algorithm
can start with the subdivision scheme (and particularly bad choices might even cause the algorithm to get stuck here).
We propose the following set of values which essentially yields an exponential search strategy:
let $m_0 \neq 0$ be a starting guess for an optimal slope and $\mu \in (0,\infty)$ an accompanying
\emph{uncertainty factor} (where $0$ would represent absolute certainty).
We choose $M_0 := [a_0, b_0] := [m_0 - \mu |m_0|,~ m_0 + \mu |m_0|]$ with $\delta_k := 2^{k+1} \mu |m_0|$.
Using these values we have that
$b_{k+1} - a_{k+1} = \delta_{k+1} = 2^{k+1} \mu |m_0| = 2 \cdot 2^{k} \mu |m_0| = 2(b_k - a_k)$
such that the interval length doubles with each expansion step.
These are also the values which we use for our following complexity analysis and numerical experiments.

Note that in principle it is possible to determine $a_0, b_0$ in such a way that the initial expansion phase isn't needed:
setting $a_0,b_0$ to be the largest / smallest slopes between two consecutive points works, and those slopes can be computed in linearithmic time.
However this is already somewhat expensive to compute and often times \emph{severely} over- / underestimates the true optimal slopes,
resulting in many more total steps than our proposed scheme in practice.

\begin{figure*}
  \centerline{
    \includegraphics[width=\textwidth]{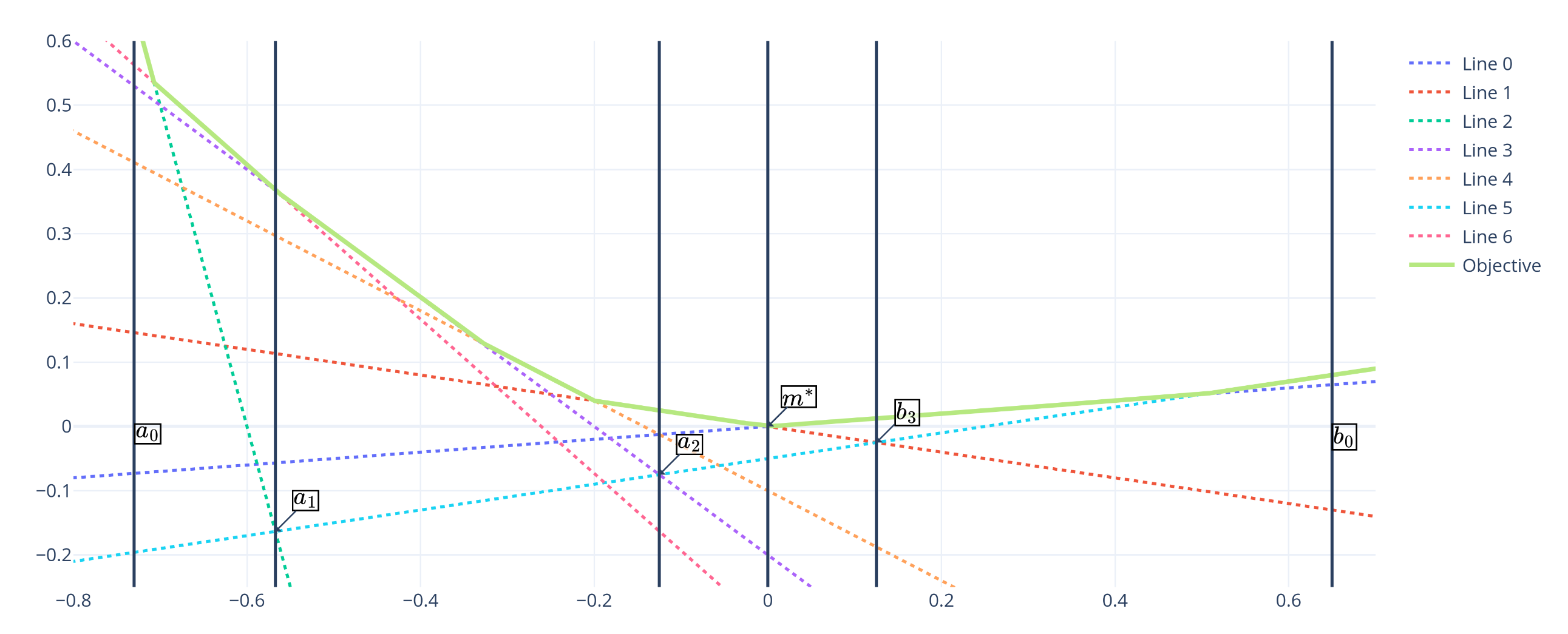}
  }
  \caption{An example showing how the interval boundaries $a_k, b_k$ evolve from one step to the next.
    Starting from an interval $[a_0,b_0]$, the approximation $J^{(1)}$ is formed by the two supporting lines at the endpoints: lines 2 and 5. The minimizer of $J^{(1)}$, found at the intersection of these two lines, becomes the new endpoint $a^{(1)}$.
    This yields a new interval $[a_1, b_1]$ with $b_1 = b_0$.
    Similarly intervals $[a_2, b_0]$ and $[a_2,b_3]$ are computed.
    At this point the approximation $J^{(3)}$ has the same
    minimizer $m^*$ as the objective and the algorithm terminates with that minimizer.
    Note that the lines 4 and 6 were ``skipped'' along the way.
    Finally note that the lines in this example do not necessarily correspond to those of an actual instance of the \L1 regression problem;
    they are instead manually chosen to qualitatively visualize the general algorithm.
  }\label{fig:subdivision}
\end{figure*}

Figure~\ref{fig:subdivision} shows a small
example of how the algorithm works.

\subsection{Complexity Analysis}\label{sec:complexity}

\begin{thm}\label{thm:main-conv}
  The proposed PALB method terminates with an exact minimizer of Problem~\eqref{eq:main-problem}
  after finitely many steps, where the number of subdivision steps is in $O(N^{4/3})$.

  Further, when initializing with $[a_0,b_0] = [m_0 - \mu |m_0|, m_0 + \mu |m_0|]$ for some $m_0 \neq 0, \mu > 0$
  and choosing $\delta_k := 2^{k+1} \mu |m_0|$,
  the method takes no more than $K$ expansion steps with $K \leq \log_2(\frac{d_{M^*}(m_0)}{\mu |m_0|} + 1) - 1$,
  where $M^* := \argmin_m J(m)$ is the set of solutions and $d_{M^*}(x) := \inf_{m \in M^*}|x-m|$ its distance function.
\end{thm}

A proof of this theorem is given in the supplementary material.
It is worth noting that the number $K$ of expansion steps is bounded by a number
that does not directly depend on the number of samples.
Moreover, in practice, it's typically possible to keep $d_{M^*}(m_0)$ ``small''
by estimating an optimal slope in some way (e.g. by solving the analogous \Lp{2} problem, or solving the \L1 problem on a subset of the data).
In the numerical experiments we will further see that in practice
the number of subdivision steps typically stays far below the given upper bound,
apparently growing logarithmically with $N$.

In the supplementary we further show a bound on the size of the expanded interval $[a_K, b_K]$
that only depends on $d_{M^*}(m_0), |m_0|$ and $\mu$.
With this bound it's possible to give alternate upper bounds on the number of subdivision
steps via the safeguarding mechanism;
notably this allows bounding the residual error after a given number of iterations
in terms of the initial error $d_{M^*}(m_0)$.

Finally it is possible to give a tighter, albeit probabilistic, bound
on the number of subdivision steps:

\begin{thm}\label{thm:main-conv-probabilistic}
  Assume the points ${\{(x_i,y_i)\}}_{i=1,...,N}$ are generated i.i.d.\
  following a distribution on $\R^2$ that admits a (Lebesgue) density function.

  Then the proposed PALB method terminates with an exact minimizer of Problem~\eqref{eq:main-problem}
  after finitely many steps, where the expected number of subdivision steps is no higher than $\frac{C N^{7/4}}{{(N/2 + 1)}^{1/4} {(N/2 - 2)}^{1/4}} \in O(N^{5/4})$ for some constant $C$ (independent of the points $(x_i,y_i)$, value of $N$ and measure).
\end{thm}

A proof of this theorem is also given in the supplementary.
It is possible to relax the assumption on the probability distribution somewhat at the cost of a minor weakening of the result:
instead of requiring a distribution with a density,
one may allow the distribution of the points to follow any Borel measure ---
i.e.\ a measure where in particular all open subsets of $\R^2$ are measurable ---
where lines have measure zero.
While most common distributions admit a density, some important examples (like points generated from an affine function with salt-and-pepper-like noise) do not.
Proving this modified Theorem proceeds completely analogous to the presented case,
just using a different result of~\cite{Leroux_2022} than the one we use in the given proof in the supplementary.
The weakening only concerns the specific value of the bound of steps, not its asymptotic behaviour.

In addition to these results it's possible to prove stronger statements for more specialized probability distributions. The strategy for this is again exactly the same one of relating the number of ``kinks'' in $J$ to the $k$-level / $k$-set problems that we use in the supplementary. A number of relevant results can be found in \cite{Barany1994}, of particular note being the expected (nonasymptotic) linear upper bound for spherically symmetric distributions on $\R^2$.

Finally: each step of the proposed method has a linear cost in the number of points $N$ of the input.
This cost is dominated by the computation of the subdifferentials $\partial J(\xi_k)$,
which principally involves partitioning ${\{(x_i, y_i)\}}_{i \in 1,...,N}$ into three parts based on whether their associated values $y_i - mx_i$
are smaller, equal to or larger than the median value
--- which can be accomplished in linear time using an introselect.
Further details on this are covered in Section~\ref{sec:impl-and-experiments} as well as Supplementary~\ref{subsec:subdiff-J}.

Combining the number of subdivision steps and per step cost yields a deterministic total complexity of $O(N^{7/3})$,
and a probabilistic one of $O(N^{9/4})$.
In our experiments in section~\ref{sec:impl-and-experiments} we will see that those bounds appear to be far from sharp:
the number of steps appears to grow logarithmically throughout all our experiments, which suggests that the PALB method might actually be in $O(N \log N)$ for an important class of problems. 

  \section{Implementation And Experiments}\label{sec:impl-and-experiments}

\subsection{Implementation Details}\label{subsec:impl}

A big appeal of the proposed method is its relatively simple implementation.
The core idea is to efficiently track the objective function $J$ and its subdifferential $\partial J$ at the interval boundaries $a_k, b_k$.
In this we principally work with the \emph{dual lines} $t_i(m) := y_i - m x_i, i=1,...,N$;
evaluating and partially sorting them based on those evaluations.
More concretely our implementation works by iteratively updating an internal state by taking expansion / subdivision steps
to generate a sequence of \emph{observable states} to support external iteration.
This proceeds until a stationary state is reached --- at which point the iteration terminates.
Alternatively a user might stop the iteration early by examining the produced states,
e.g.\ when the observable states already show a small enough objective value for their purposes or when a maximum number of iterations is reached.

The iterator principally consists of
a buffer holding all the datapoints (implemented as an array), a second buffer holding pairs of dual lines and associated values (again using a contiguous array), 
some auxiliary data (how many steps were taken, whether or not subdivision has started etc.), and two \emph{states} containing information about the objective function at the respectively left and right interval boundaries. We describe these next.

The two boundary states each consist of a slope $m$, the median line at that slope (i.e.\ some line out of all dual lines $t_i$ which has median value at $m$), the subdifferential and value of the objective function at $m$.

The step function of the iterator then directly implements Algorithm~\ref{alg:main} fairly directly,
with the central difference being that instead of recursing explicitly, the internal iterator state is updated.
The necessary computations can all be carried out by evaluating the dual lines at some slope, writing the resulting values into the buffer on the iterator, and then partitioning the full buffer using two introselects (to obtain the three-way partition into values larger, equal and smaller than the reference median). We use Rust's built-in \texttt{slice::select\_nth\_unstable\_by\_key} function for these partitions\footnote{Note that when implementing the method in other languages, good care has to be taken at this point to not negatively impact the theoretical guarantees or practical performance: certain implementations of the same basic partitioning functionality (e.g. C++'s \texttt{std::nth\_element}) exhibit a worse asymptotic complexity, and others (notably Python's \texttt{numpy.partition}) don't work in-place and as such incur a full copy of the data each time.}.

Our implementation by-default uses an uncertainty of $\mu = 0.01$ and computes the starting guess $m_0$ for the optimal slope by solving the analogous \Lp{2} regression problem. For particularly small data a cheaper to compute starting guess can be employed; we empirically found that using the line passing through the first and last data point works well for $N \leq 100$ (we don't assume the data to be sorted, but if it turns out to be then the first and last points may provide a better estimate than just taking the first few; however it of course is a heuristic either way).
Further in each subdivision step, the safeguarding intervals $S_k$ are chosen to be $[a_k + \varepsilon_k, b_k - \varepsilon_k]$ with $\varepsilon_k := 0.01 \cdot |b_k - a_k|$.

We implement multiple mechanisms for avoiding numerical issues and improving accuracy:
for one we use the Kahan-Babuška-Neumaier compensated summation \cite{Neumaier1974} for most sums (in particular we use a single such compensated summation to evaluate the difference of the sums of the $x$ values from above and below the median line. This is a point where catastrophic cancellation can occur otherwise).
When computing the points of intersection of the lines $l_a(m) = s_a (m - a) + f_a, l_b(m) = s_b (m - b) + f_b$ we do so in transformed coordinates where the midpoints $\tfrac{a+b}{2}$ is at zero, and then shift the result back to the original coordinates.
As a very first step in our algorithm we also (optionally) translate the input data such that its centroid is at $(0,0)$ and scale it such that it is contained in the square $[-1,1]^2$ --- the solution to the original problem is obtained by applying the inverse scaling and translation. This is justified and accomplished as follows:

Suppose we express data $\{(x_i, y_i)\}_{i=1,...,N}$ in terms of some affinely transformed data $x_i = \sigma_x x_i' + \tau_x, y_i = \sigma_y y_i' + \tau_y$ and assume $\sigma_x, \sigma_y \neq 0$ such that the transformation is invertible. Then it holds that
\begin{align*}
  \sum_{i=1}^N |m x_i + t - y_i| &= \sum_{i=1}^N |m (\sigma_x x_i' + \tau_x) + t - (\sigma_y y_i' + \tau_y)|
  \\ &=  |\sigma_y| \sum_{i=1}^N |\underbrace{(m \frac{\sigma_x}{\sigma_y})}_{=:m'} x_i' + \underbrace{\frac{t + m\tau_x - \tau_y}{\sigma_y}}_{=: t'} - y_i'|
\end{align*}
Hence $m,t$ solves the \L1 problem on the original data iff $m', t'$ solves the \L1 problem on the transformed data.

Finally, we have added safeguards to prevent infinite iteration by capping the number of iterations to $15\log_{10}(N) + 300$ (with the log being an integer logarithm) and terminating once $|b_k - a_k| < 10^{-15}$ during bisection.
The iteration-bound is an empirical one:
as our experiments show the actual growth of the necessary number of steps is approximately $5\log_{10}(N)$,
including a ``safety factor'' of 3 we get a prefactor of 15,
and the 300 make it so that we rarely actually hit this boundary (cf.\ Figure~\ref{fig:steps-both}).

For further details we refer to our reference implementation
at~\url{https://github.com/SV-97/piecewise-affine-lower-bounding}. The method described in this paper corresponds to \texttt{palb} version \texttt{0.1.2}.

\subsection{Experimental Setup}\label{subsec:experiments}

\subsubsection{List Of Methods}

We selected the following methods to compare the proposed algorithm with:

\paragraph*{L1pack --- Barrodale-Roberts Algorithm}
The Barrodale-Roberts algorithm is a quite long-established method for the \L1 regression problem~\cite{BarrodaleRoberts_1974} that has seen some improvements over time (cf.~\cite{Bloomfield_1983}). At its core this is a bespoke simplex method that exploits special structures inherent to the \L1 problem. A C / Fortran implementation with R interface is available as part of the L1pack library.

\paragraph*{IBM CPLEX --- Simplex Method}
There are multiple possible LP formulations for the \L1 regression problem that are known to have drastically different runtime characteristics.
We will use the following dual formulation:
\begin{align}\label{eq:dual-lp}
  \max_{d \in \R^N} d^T y \quad \text{s.t.}~ d^T x = 0,\ d^T 1_{N} = 0,\ -1 \leq d \leq 1
\end{align}
This is a standard formulation, cf. \cite{armstrong1982dual}.
It is folklore that this formulation performs well in practice;
we also found this confirmed in our testing.
The optimal slopes and intercepts are obtained as the Lagrange multipliers of the two equality
constraints associated to a solution of this dual problem.
We use the standard parameters for CPLEX which, according to the logs (these were not enabled for the actual benchmark runs so as to not influence the results, but rather for a separate run where we looked into what CPLEX was actually doing), results in using CPLEX's parallel strategy: it starts the various algorithms that it implements (Dual simplex, primal simplex, Barrier and Sifting methods) in parallel, and returns the result of the first method that terminates.
A cursory look through the logs suggests that for the vast majority of problems, the method that finished first was the dual simplex method (which is also why we list CPLEX as a simplex method), but in rare instances the Barrier method finishes first.

\paragraph*{Clarabel --- Interior Point Method}
We also solve the above LP formulation using Clarabel~\cite{Clarabel_2024},
a recently published library that implements a general-purpose interior-point solver for conic programs with quadratic objectives.
Notably it is distributed as a default solver with the well established Python library CVXPY
for convex optimization, where it is in particular used for linear programs.
Clarabel has implementations in both Rust and Julia; we use the Rust version.

\paragraph*{HiGHS --- Parallel Simplex Method}
HiGHS~\cite{HiGHS_2018} is a high-performance solver that in particular implements parallel algorithms for solving LPs.
It is implemented in C++.
Similarly to CPLEX and Clarabel we apply HiGHS to the dual LP formulation~\eqref{eq:dual-lp}.

\paragraph*{Statsmodels --- Iteratively Reweighted Least Squares}

Statsmodels \cite{seabold2010statsmodels} is a standard Python package implementing various statistical models and tests.
In particular it implements an inexact method, IRLS, for solving quantile regression problems of which the L1 regression
problem is a particular instance.
This method is implemented in Python, but internally uses the C/Fortran-backed numpy for acceleration.

Automatic algorithm and hyperparameter selection was used for each solver.
For specifics on the versions and APIs used for our experiments please see Table~\ref{tab:software-versions}.

\begin{table}
  \begin{center}
    \caption{
      Versions of the software used in our experiments.
      Binding Language refers to the language bindings used for implementing the associated experiments, whereas Implementation Language refers to the language in which the actual solver is implemented in.
      Binding Package refers to the precise API used.
      Notably for CPLEX we did not use the higher-level ``modeling'' API.
      We also tested CPLEX with its Python API; this yielded slightly worse times for smaller sample sizes but no appreciable difference on larger ones.
      We used Python \texttt{3.13.7}, Rust \texttt{1.88.0} and R \texttt{4.3.3} when running our experiments. 
    }
    \setlength{\tabcolsep}{3pt}
    \begin{tabular}{p{35pt} p{65pt} p{35pt} p{45pt} c}
  Name & Implementation Language & Binding Language & Binding Package & Version \\
  \hline
  Clarabel & Rust & Rust &\texttt{good\_lp} & \texttt{0.11.1} \\
  CPLEX & C & Rust &\texttt{good\_lp} & \texttt{22.1.2.0} \\
  HiGHS & C++ & Rust& \texttt{good\_lp} & \texttt{1.12.0} \\
  L1pack & Fortran, C & R &\texttt{l1pack} & \texttt{0.60} \\
  Statsmodels & Python (w/ Numpy) & Python & \texttt{statsmodels} & \texttt{0.14.5} \\
  PALB (Ours) & Rust & Rust & \texttt{palb} & \texttt{0.1.2} \\
\end{tabular}\label{tab:software-versions}
  \end{center}
\end{table}

\subsubsection{List Of Experiments}

We discuss four experiments we use to compare the proposed method with the state-of-the-art methods.
Three of the experiments use synthetic data and investigate the effects of various properties of the data on the solver performance,
and one experiment uses real data.

\paragraph{Linear Groundtruth}
We generate $1\,000$ random lines $g$ of the form $g(x) = m x + t$ such that
$g(0), g(1)$ are uniformly distributed in $[0,1]$.
For $N$ ranging from $10$ to $10\,000\,000$ we consider each such line, generate points
${\{x_i\}}_{i=1,...,N}, x_i \sim \Uniform([0,1])$ and obtain noisy samples
${\{y_i = g(x_i) + \epsilon_i \}}_{i = 1,...,N},\ \epsilon_i \sim \Laplace(0, 0.1) + \Uniform([-0.05, 0.05])$
for those points.
We then measure the runtime of every considered method on ${\{(x_i, y_i)\}}_i$.

\paragraph{Non-linear Groundtruth}
This experiment is very similar to the one before, however with (typically) highly nonlinear ground truths.
Note that the lines $g$ from before can be written as $g(x) = \alpha x + \beta (1-x)$
with $\alpha, \beta \sim \Uniform([0,1])$; i.e.\ they have have uniformly distributed coefficients
in the Bernstein basis.
Instead of lines we now consider polynomials of degree 5 that are uniformly distributed in the sense
that all their coefficients in the Bernstein basis are again $\Uniform([0, 1])$ distributed.
In particular these polynomials are maps $[0,1] \to [0,1]$.
The noise model is the same as before: a mix of uniform and Laplace noise.
The purpose of this experiment is to investigate what happens when the model of a line does not match the underlying ground truth of the given data.

This is interesting because it influences the distribution of the kinks in the objective function as well as certain
approximate solution schemes (that may be used to seed an exact solver):
without any noise and for an affine-linear ground truth,
the dual lines corresponding to a sample would all intersect in a single point.
Adding noise then causes these intersections to deviate from this point somewhat but we can still
expect them to be in the rough vicinity of the original point of intersection.
But for general input data this distribution can not be expected to hold.
To test this we consider an intentionally nonlinear ground truth.

\begin{figure*}
  \centerline{
    \includegraphics[width=\textwidth]{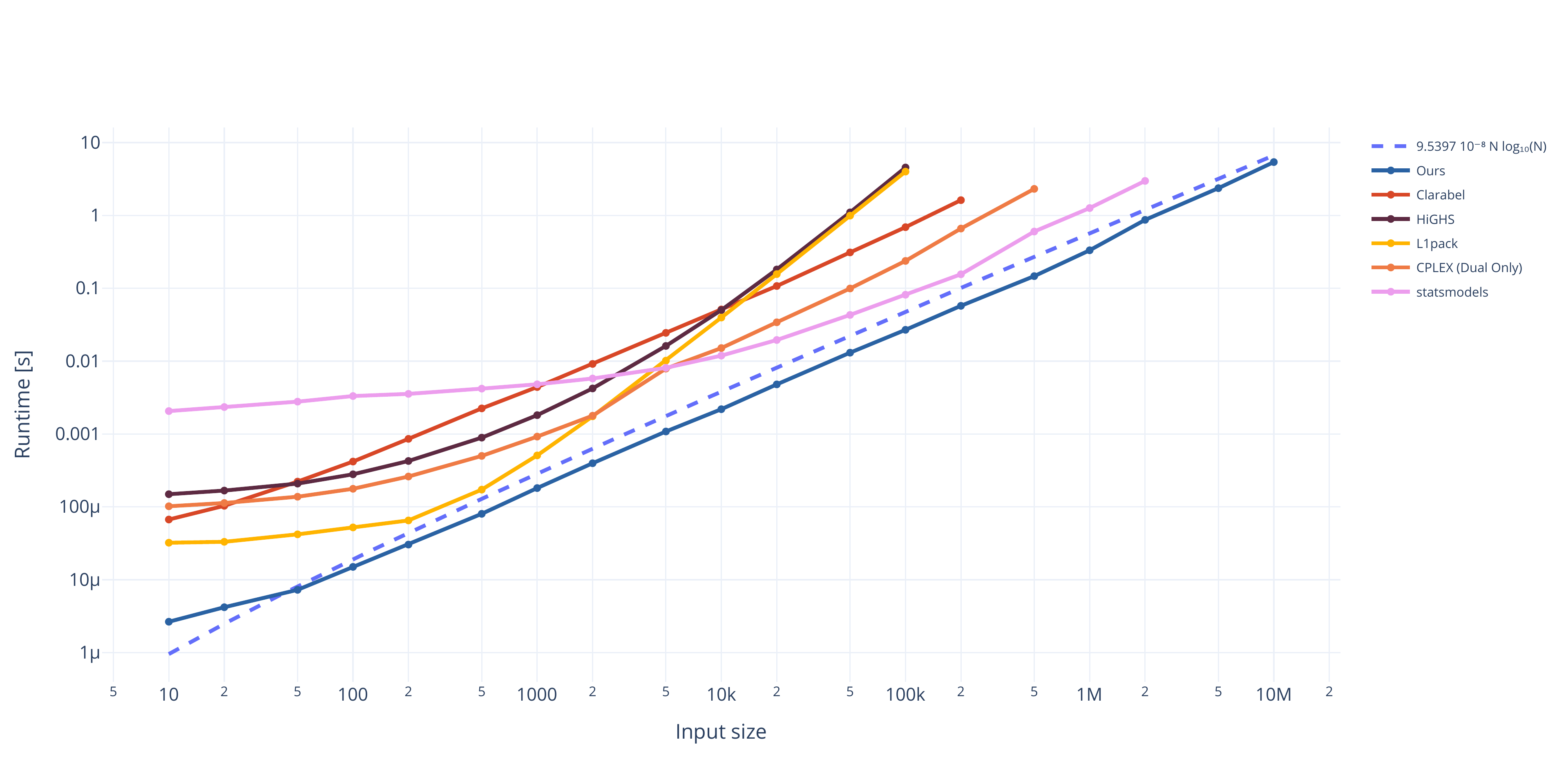}
  }
  \caption{Median runtime results for the experiments on synthetic data.}\label{fig:runtimes-synthetic}
\end{figure*}

\paragraph{Linear Groundtruth with Severe Outliers}
We again consider lines as in Experiment 1, however we now consider a mixture of gaussian and
Cauchy noise: we generate the points ${\{x_i\}}_{i=1,...,N}, x_i \sim U([0,1])$ and
sample $g$ to obtain ${\{y_i + \epsilon_i\}}_i$ where
\begin{align*}
  G_i \sim \Laplace(0, b), \quad &C_i \sim \Cauchy(0, \gamma),
  \\ \delta_i \sim \Bernoulli(1-p), \quad&\epsilon_i := \delta_i G_i + (1 - \delta_i) C_i
\end{align*}
with $b := 0.01, \gamma := 0.5, p := 0.05$.
So in effect approximately 5\% of the samples will be outliers.

\vspace*{1em}
All these experiments consider data with ground-truth data $x_i, y_i \in [0,1]$.
This is done to simplify noise models and justified by the transformation properties outlined in the implementation section.
One can always arrange for all data to be in ${[0,1]}^2$ and simply transform the solution afterwards.

Note that the probability distributions used in these experiments all satisfy the conditions of our complexity result Theorem~\ref{thm:main-conv-probabilistic},
since they all admit densities.

\paragraph{ISD Temperature Data}\label{par:experiment-4}

Finally, we consider a data sample of the NOAA's \emph{Integrated Surface Database} (ISD).
The ISD is best described by quoting the official website~\cite{NOAA:ISD}: ``The Integrated Surface Database (ISD) is a global database that consists of hourly and synoptic surface observations compiled from numerous sources into a single common ASCII format and common data model. ISD integrates data from more than 100 original data sources, including numerous data formats that were key-entered from paper forms during the 1950s–1970s time frame. ISD includes numerous parameters such as wind speed and direction, wind gust, temperature, dew point, cloud data, sea level pressure, altimeter setting, station pressure, present weather, visibility, precipitation amounts for various time periods, snow depth, and various other elements as observed by each station.''

We randomly selected 681 stations and downloaded their available ISD data. We aggregated all that data into one timeseries per station. The columns we work with are the measurement times and dry-bulb temperature in \SI{}{\celsius}.

The L1-regression problem on this data is to regress the dry-bulb temperature in \SI{}{\celsius} against time for each series.
For this we mapped the given datetime values to real numbers as follows: we consider a coordinate system centered at the first of January 1950 (UTC), scaled such that each year (assumed to have \SI{31557600}{\second}) corresponds to 1 unit of time.

During processing we dropped any rows with missing data (\emph{nulls}), i.e. those rows of data where no actual dry-bulb measurement was recorded in the ISD. We furthermore dropped any stations where the number of available measurements (after dropping nulls) was less than 10.

This resulted in 645 timeseries with lengths ranging from 10 to 1\,977\,748 samples.
For the purposes of evaluating L1-regression methods this dataset has multiple interesting properties: it often times contains very large gaps in time (years where no data was recorded by a station), can be locally highly nonlinear (e.g. because of periodic temperature differences across a day of measurements), and can have substantial qualitative differences over time (e.g. changes to how often data is recorded).  Moreover the drastically different lengths of the various series means that to efficiently process the whole dataset a method must be fast for both small as well as large input sizes.

\begin{figure*}
  \begin{subfigure}{.5\textwidth}
    \centerline{
      \includegraphics[width=\columnwidth]{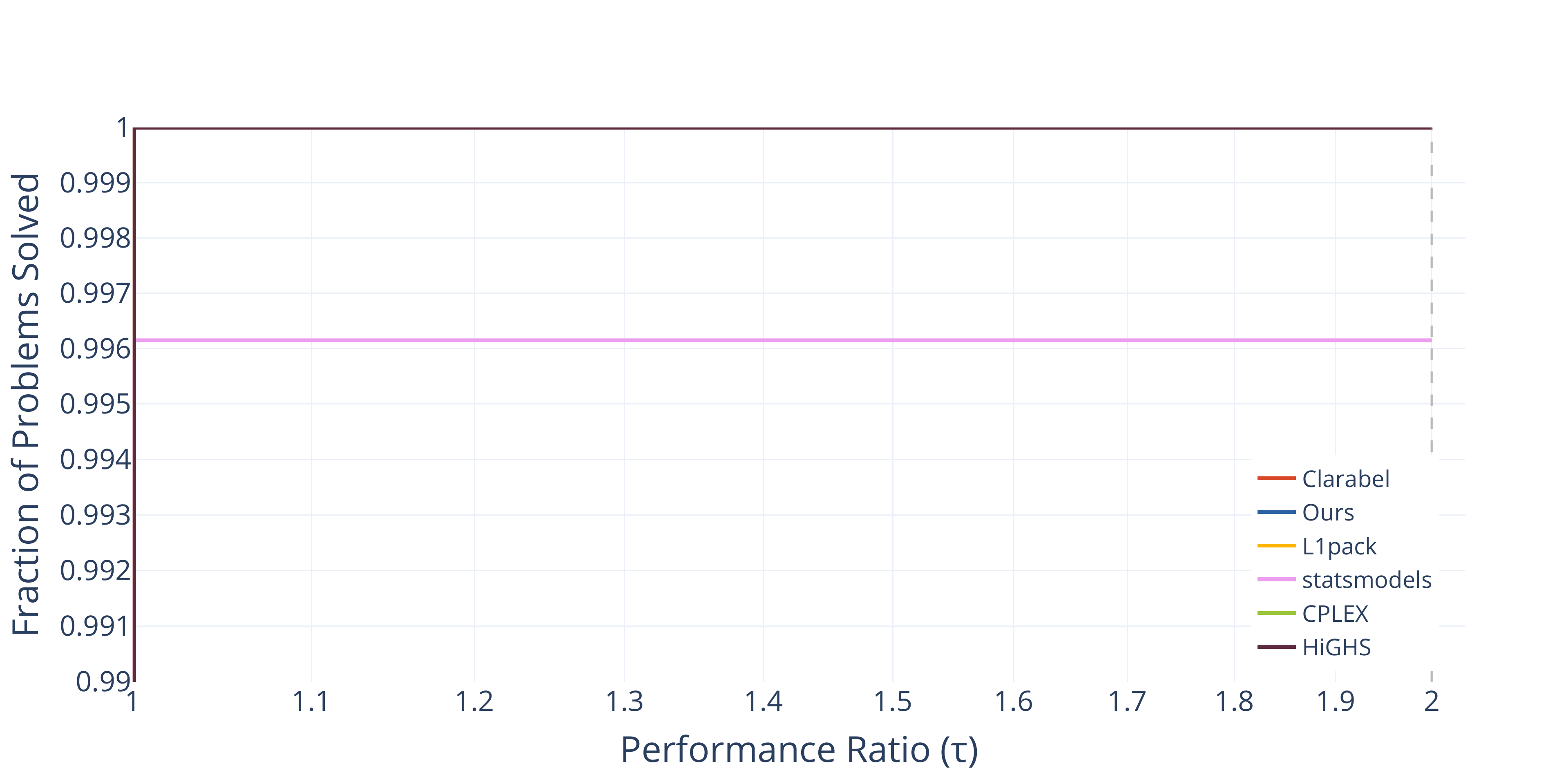}
    }
    \caption{Objective value at termination.
    }\label{fig:perfprof-obj-synth}
  \end{subfigure}
  \begin{subfigure}{.5\textwidth}
    \centerline{
      \includegraphics[width=\columnwidth]{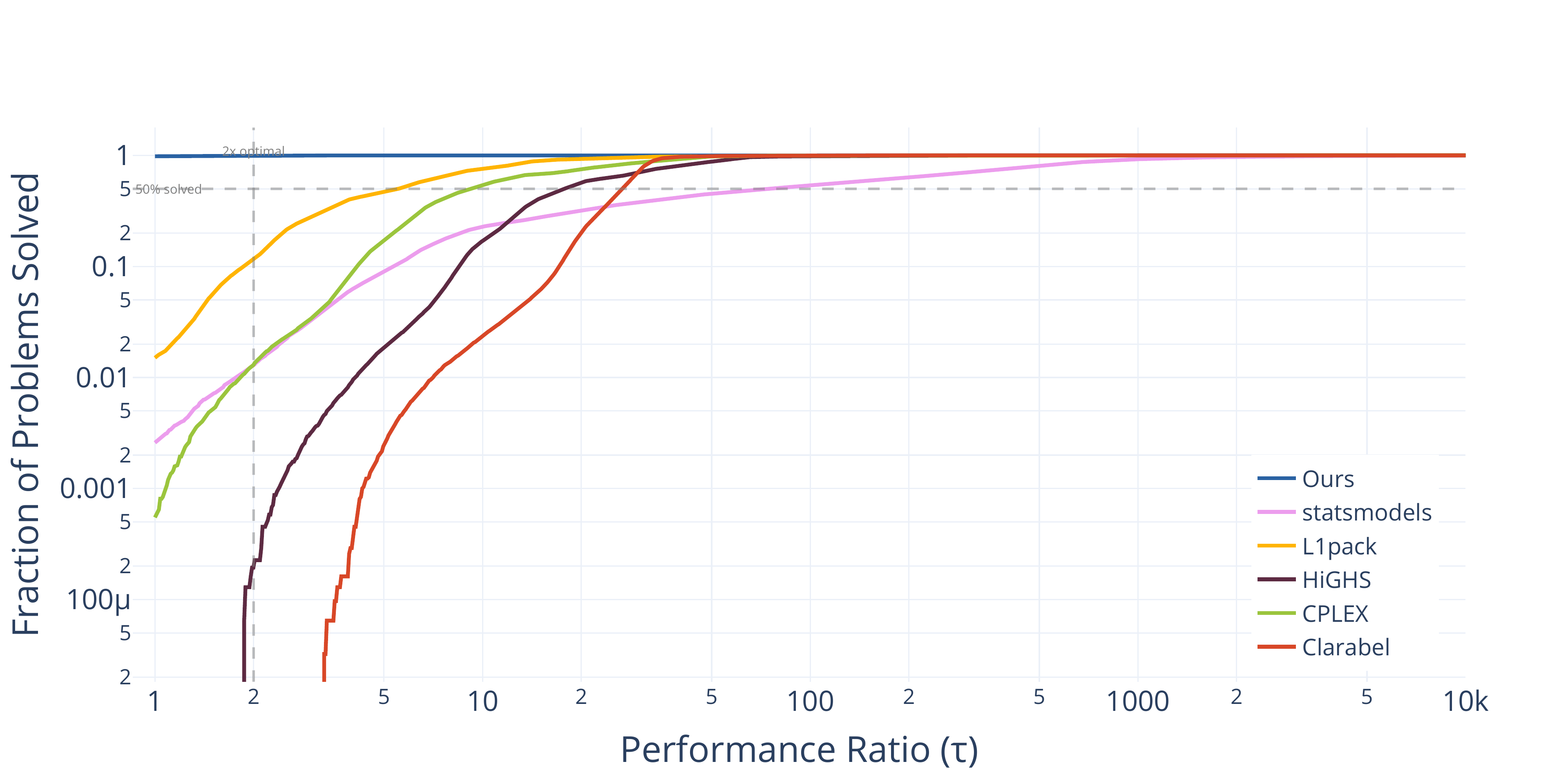}
    }
    \caption{Runtime.
    }\label{fig:perfprof-time-synth}
  \end{subfigure}
  \caption{
    The performance profiles for the synthetic data.
  }
\end{figure*}

\begin{figure*}
  \begin{subfigure}{.5\textwidth}
    \centerline{
      \includegraphics[width=\columnwidth]{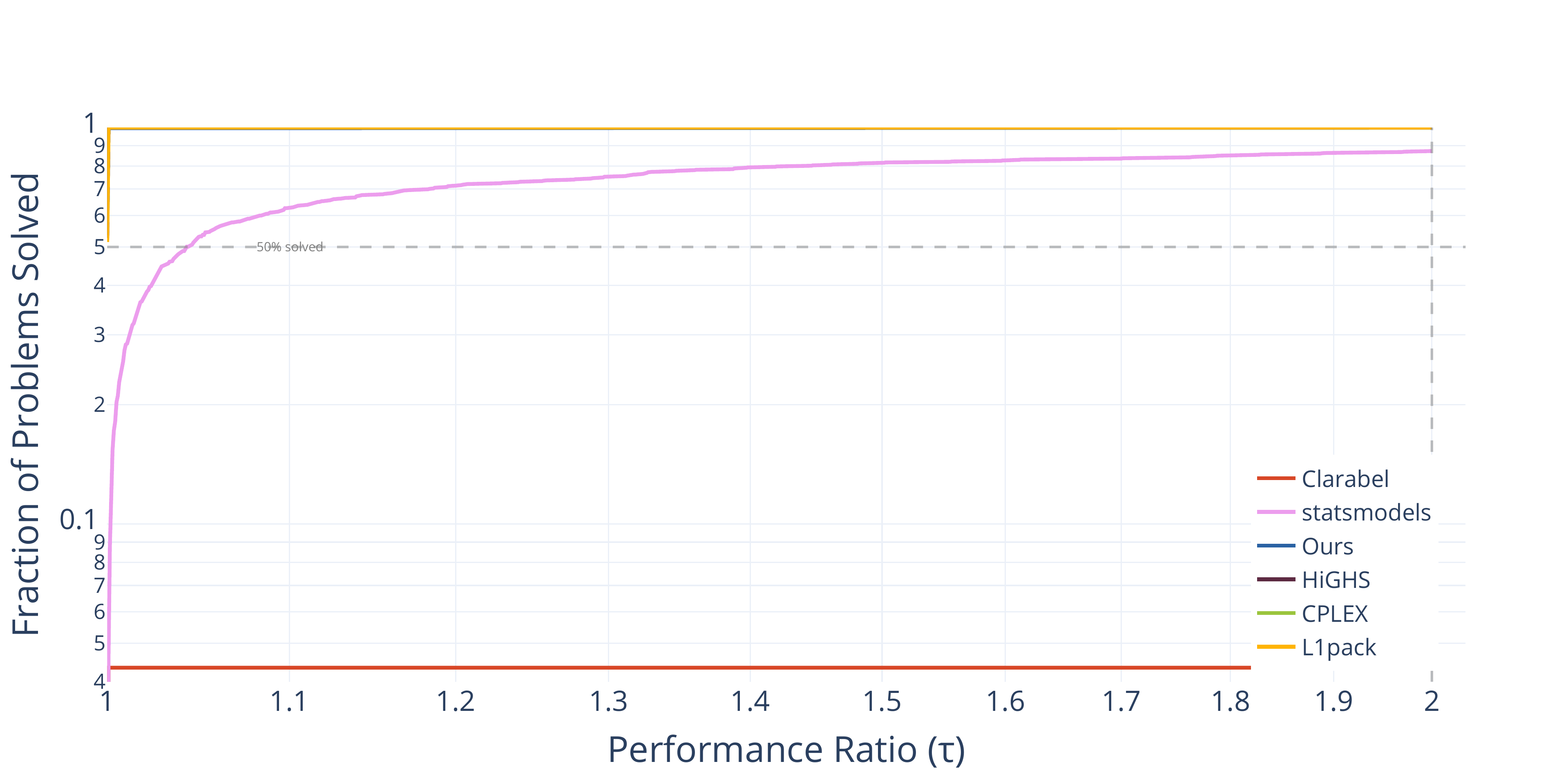}
    }
    \caption{Objective value at termination.
    }\label{fig:perfprof-obj-real}
  \end{subfigure}
  \begin{subfigure}{.5\textwidth}
    \centerline{
      \includegraphics[width=\columnwidth]{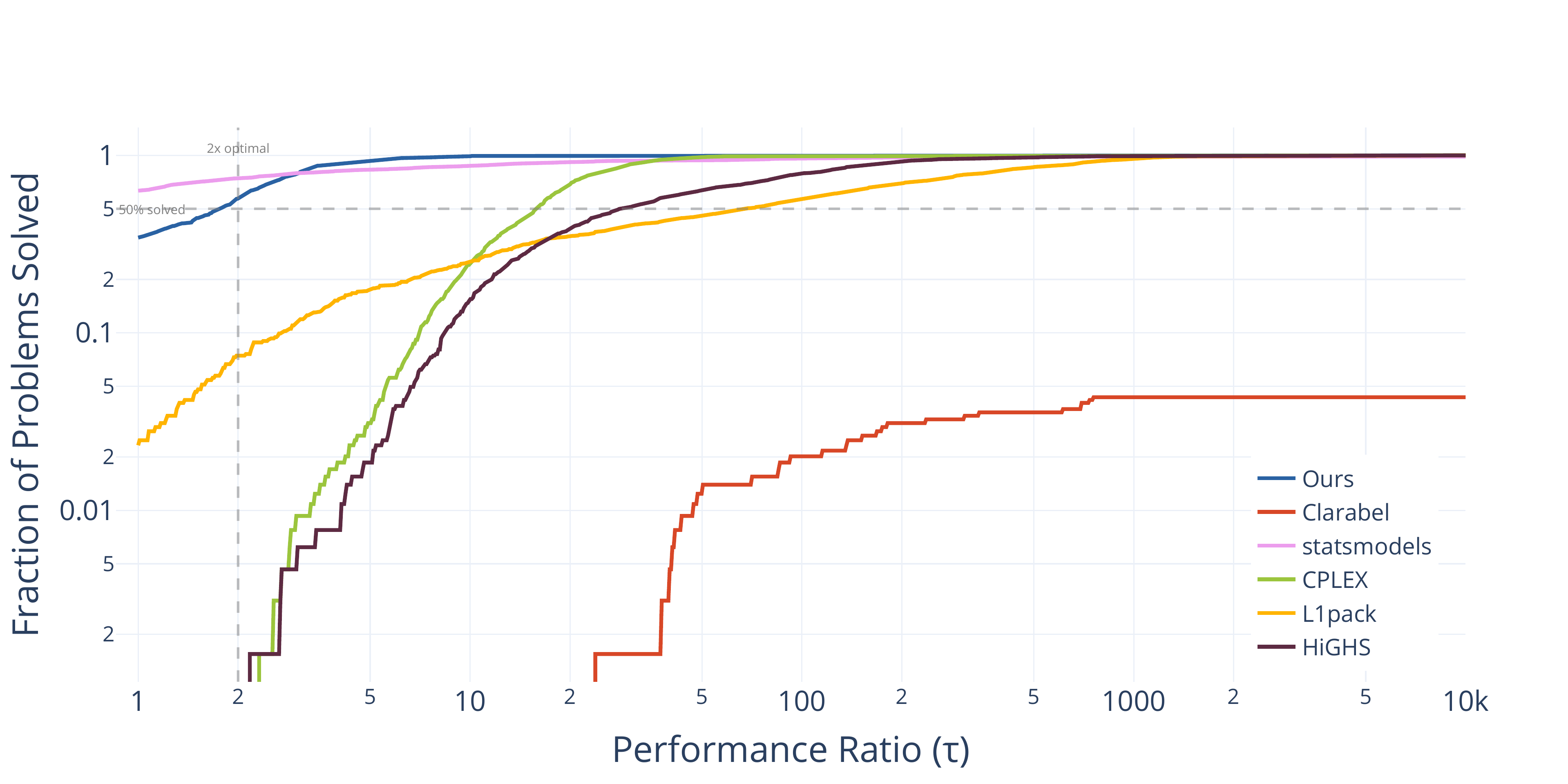}
    }
    \caption{Runtime.
    }\label{fig:perfprof-time-real}
  \end{subfigure}
  \caption{
    The performance profiles for the real data.
  }
\end{figure*}

\paragraph{Hardware Setup}

All benchmark data was obtained on a recent desktop computer with an AMD Ryzen 9 5900X processor clocked at \SI{4.6}{\giga\hertz}, \SI{64}{\gibi\byte} of DDR4 RAM at \SI{3200}{\text{MT/s}}. Processor-specific optimizations (i.e. \texttt{-Ctarget-cpu=native}) like autovectorization, as well as link-time optimizations (\texttt{lto=true}) were enabled for all our Rust compilations. Furthermore we used the high-performance \texttt{mimalloc} memory allocator throughout the Rust-based experiments. This ensures that none of the native methods were severely limited by their memory management (it should be reemphasized that this isn't a potential problem for our method since we don't allocate much to begin with).

\subsection{Performance Profiles}

We aggregate the results into two groups -- synthetic and real data -- and compute performance profiles \cite{Clarabel_2024, DolanMore2002} for each one.
In a nutshell, the value $\rho_s(\tau)$ of the performance profile function $\rho_s : \R_{\geq 1} \to [0,1]$ gives the probability that the score (i.e.\ objective function value or runtime) of solver $s$ on a randomly chosen problem from a fixed collection of problems, is at least within a factor $\tau$ of the best value among all the solvers from a reference group.
So it is desirable for a solver to have a large value of $\rho_s(\tau)$ for a comparatively small value of $\tau$, ideally even $\rho_s(1) = 1$.
If $\rho_s$ never attains the value $1$ for a given solver, this means that it failed to solve some problems from the problem collection under consideration.
In supplementary~\ref{appendix:sec:perfprof} we give a more detailed introduction to these performance profiles.
There we also discuss associated methodological choices.
When discussing the experimental results we write $\rho^{\mathrm{(obj)}}$ and $\rho^{\mathrm{(time)}}$ to specifically refer to the profiles associated to the objective values and runtime data respectively.

\subsection{Numerical Results}

\paragraph{Runtimes on Synthetic Data}

Figure~\ref{fig:runtimes-synthetic} shows the median runtimes for the various methods on the synthetic input data
(aggregated over all three experiments since the results are very similar across these experiments).

Whenever a method did not produce a solution (i.e. due to nonconvergence) on some sample we set its corresponding runtime to $+\infty$.
In this case the mean is no longer well defined (and hence omitted), but the median still is.
In total there were 119 nonconvergences of IRLS across the $32\,100$ test datasets, all occuring in the range from $N=10$ to $N=2\,000$.
\quad
Note that both axes are scaled logarithmically so that an apparent constant distance between two traces
indicates that they are multiples of one-another rather than actual translates.
In the case of synethetic data all the methods were ran up to the sample size where they first exceeded a median runtime of $1\si{\second}$;
past this point the experiment runtimes got prohibitively long on the hardware we used.

\begin{figure*}
  \begin{subfigure}{.5\textwidth}
    \centerline{
      \includegraphics[width=\textwidth]{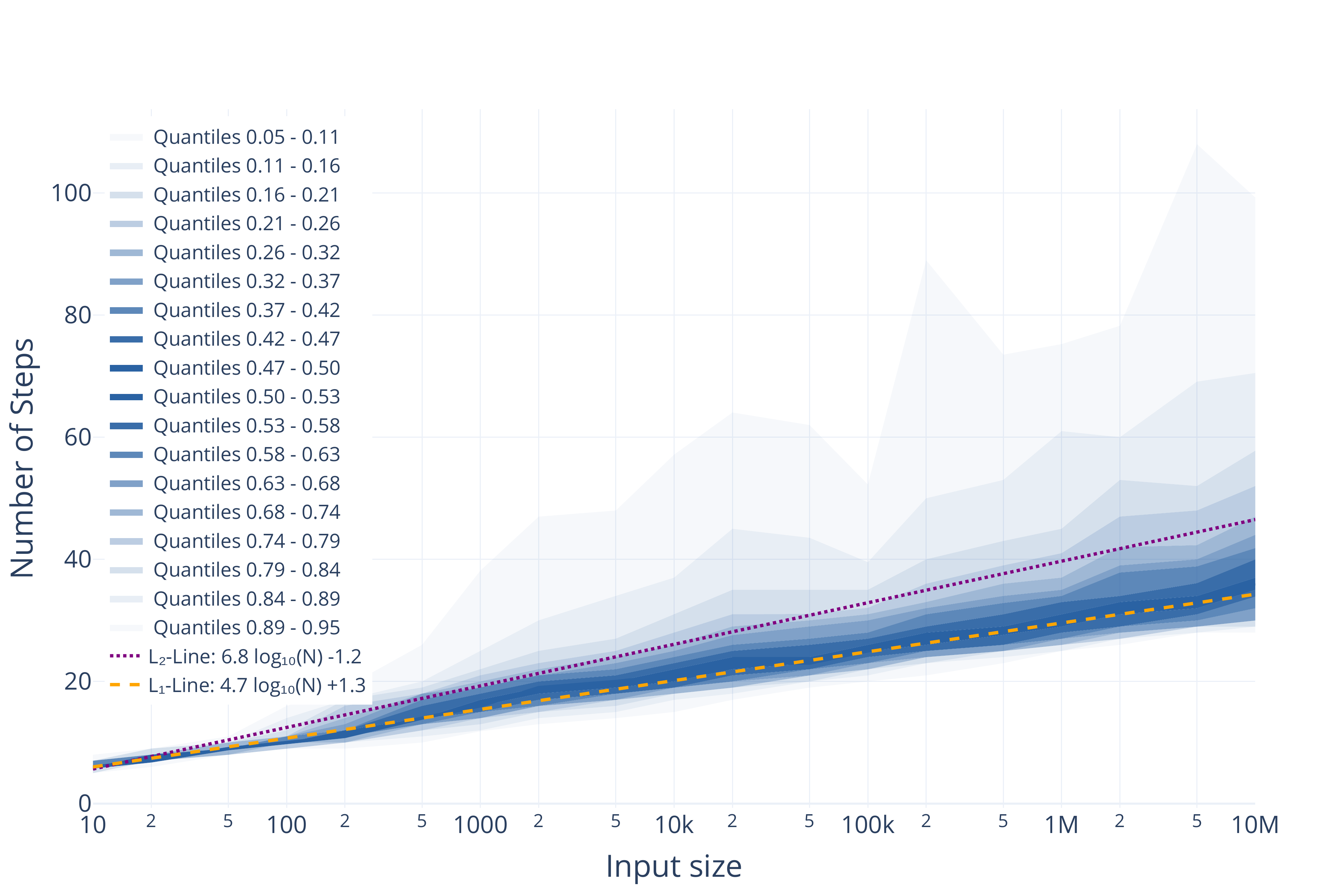}
    }
    \caption{
      Synthetic data.
    }\label{fig:steps-synth}
  \end{subfigure}
  \begin{subfigure}{.5\textwidth}
    \centerline{
      \includegraphics[width=\columnwidth]{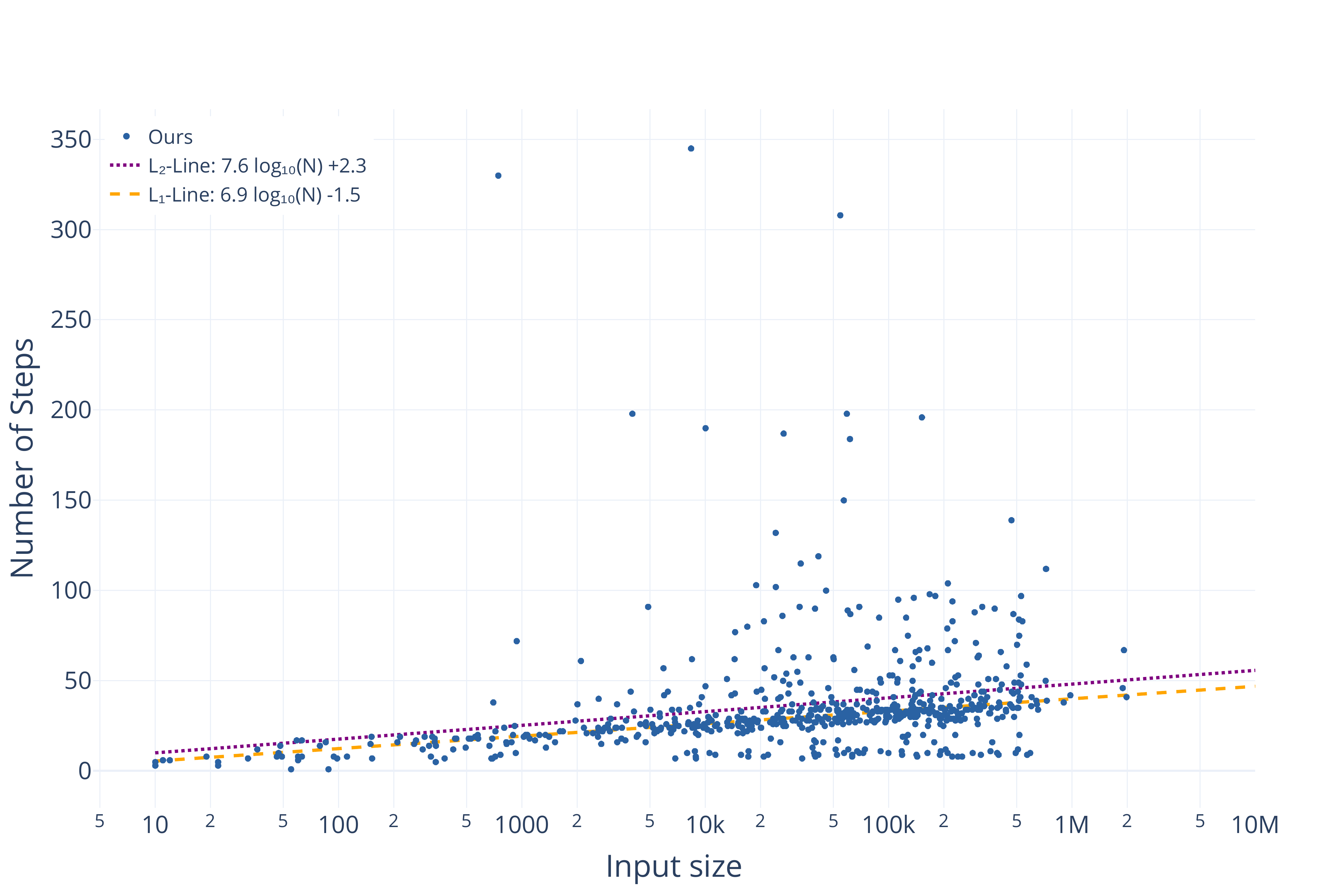}
    }
    \caption{
       Real data.
    }\label{fig:steps-real}
  \end{subfigure}
  \caption{
    The number of steps our method took to produce a solution on synthetic and real data.
    (a) The indicated regions show the distribution of the steps across all realizations. Each region is bounded by the indicated $p$-quantiles.
    (b) Each point indicates the number of steps taken for one timeseries.
    (The synthetic and real data are visualized differently because plotting distinct datapoints for every sample in the synthetic data looks rather cluttered, while there aren't enough samples per input size to produce a meaningful quantile plot in the case of real data. The dashed and dotted lines show \Lp{1} and \Lp{2} lines fitted to the measurements. The \Lp{2} fit is accomplished using numpy's \texttt{np.linalg.lstsq} method, while the \Lp{1} fit is due to our method. There are some points in the samples that can be considered outliers --- these can be seen to throw the \Lp{2} lines off from what one might intuitively expect.) 
    }
    \label{fig:steps-both}
\end{figure*}

Going through all methods in order of how fast the are at $N=10$:
Our proposed method works well throughout; quickly settling in to its final trend.
\quad
The Barrodale-Roberts algorithm of L1pack performs well initially but can be seen to scale roughly quadratically eventually, causing most other methods to outpace it for large $N$.
\quad
The interior point method of Clarabel also performs well for very low numbers of samples
and while it is quickly overtaken by most other methods around $N=100$,
it eventually, at approximately $N=10\,000$, outperforms L1pack and HiGHS due to its superior scaling properties.
\quad
The commercial simplex method of CPLEX is roughly in the middle for most input sizes,
performing comparably well to L1pack between $N=2\,000$ and $N=5\,000$.
Of particular note is CPLEX's apparent ``kink'' at $N=5\,000$ which is likely a result
of some internal parameters being chosen differently at that point.
\quad
The final LP-based method HiGHS starts out just slightly slower than Clarabel, but eventually scales roughly quadratically just like L1pack. It seems to be ill-suited for our specific problem.
In our experiments we also evaluated the \texttt{QuantileRegressor} of \texttt{scikit-learn} which internally uses HiGHS by-default
--- aside from being slightly slower than bare HiGHS (likely due to the overhead incurred by the Python API)
we observed similar results for this method.
\quad
Finally there is the IRLS (iteratively reweighted least squares) method implemented by statsmodels.
This is the only non-exact method in our tests.
Note the method failed to converge for 119 out of the total of $32\,100$ inputs, all failures occuring in the range up to $N=2\,000$.
We handled these nonterminating cases as an infinite runtime, so they are accounted for in the plotted median line.
The method exhibits an approximately constant runtime profile up to about $N=2\,000$.
This could be explained by the overhead of its Python implementation dominating the actual compute time, or some other internal overhead.
Past this point the scaling starts ramping up,
but it is still sufficiently low for the IRLS method to settle in between ours and the other exact methods.
Remarkably, despite its inexact nature it doesn't outperform our method locally
and also doesn't seem to outpace it as the input size grows.
\quad
The indicated log-linear trend-line was fitted to the times of our method using the nonlinear least-squares method implemented by \texttt{scipy}.

\paragraph{Performance profiles on synthetic data}

The performance profile for the objective function on synthetic data shows no major deviations between the different methods; for completeness it is given in Figure~\ref{fig:perfprof-obj-synth}.
It shows that our method has identical results to Clarabel and HiGHS, and that the various methods are effecively equivalent in this case. The only standout is statsmodels which didn't solve all problems in the test set.

The performance profile for the runtime shown in Figure \ref{fig:perfprof-time-synth} confirms what Figure~\ref{fig:runtimes-synthetic} already suggested: our method almost universally outperforms the other methods to a quite substantial margin, with even the second best method, L1pack, only managing to solve approximately $50\%$ of problems within approximately $6.5$ times the time of the best method on those problems.

\paragraph{Performance profiles on real data}

The performance profiles in Figures~\ref{fig:perfprof-time-real} and~\ref{fig:perfprof-obj-real}, obtained from the ISD data experiments, visualize the practical advantages of our method for real world data analyses:
Compared to other exact solvers, our proposed method offers a significant speedup while retaining optimal objective values. For instance, CPLEX and L1pack reach a value of $\rho^{\mathrm{(time)}}(\tau) = 0.5$ only at approximately $\tau = 15.89$ and $\tau = 66.35$ respectively, while ours does so at $\tau \approx 1.8$.

In this test the inexactness of IRLS becomes quite evident: in the objective value profile it reaches $50\%$ solved at approximately $\tau = 1.04$, but then has a rather long ``tail'' (i.e.\ it produces comparatively bad objective values for a subset of the problems) giving for example $\rho^{\mathrm{(obj)}}_\text{statsmodels}(1.3) \approx 0.749$ and (not shown in the figure) $\rho^{\mathrm{(obj)}}_\text{statsmodels}(2.32) \approx 0.9$.
In the runtime profile the IRLS method can be seen to be faster than our method on a subset of the problems (for low $\tau$), however it again suffers from a very long tail as $\tau$ increases. This indicates that it takes a substantial time to solve some of the problems. Because of this our method surpasses it at approximately $\rho^{\mathrm{(time)}}(3) \approx 0.78$.
Moreover, IRLS actually doesn't converge for 10 out of the 645 problems in this test set.

Finally, Clarabel struggled significantly with this dataset, failing to produce a solution in 617 out of 645 cases due to (as reported by its \emph{verbose} output; which we temporarily enabled for troubleshooting purposes) ``insufficient progress'' when using default settings.
Although we attempted manual parameter tuning (specifically tweaking equilibration, presolve, tolerances, and regularization settings), we were unable to reduce the number of nonterminations below 467. Consequently, and to keep the data more easily comparable, we reverted to default settings for the reported experiments.

\paragraph{Number of needed iterations for our method}

Along with the runtime data we recorded the total number of steps taken by our
method until termination for each test dataset.
These numbers are visualized in Figures~\ref{fig:steps-synth} and~\ref{fig:steps-real}.
\quad
For the synthetic data we compute quantiles of the step numbers across all these realizations grouped by the number of samples.
The shaded regions indicate these quantiles.
\quad
In both cases we see that the distribution opens up somewhat as $N$ increases, but the primary mass always stays located along a logarithmic trend curve (fitted using numpy's least squares method after removal of outliers by regressing the runtimes against $\log_{10}(N)$).
Note that the x-axis is logarithmic so that the apparent linear growth really indicates a logarithmic
growth of the number of steps as a function of the number of points in the sample.

  \section{Conclusion}

This work presented the Piecewise Affine Lower-Bounding (PALB) method, an exact and computationally efficient algorithm for LAD line fitting. PALB reformulates the LAD objective as a convex, piecewise-linear function of the slope and iteratively refines affine lower bounds from subgradient information to locate the global minimizer with finite termination.

Theoretical analysis confirms correctness and relates the number of required iterations to the $k$-level problem in combinatorial geometry, providing an $O(N^{\frac{4}{3}})$ upper bound on the number of iterations or an $O(N^{\frac{5}{4}})$ bound in expectation under additional mild assumptions on the data distribution.
Empirical results indicate that the iteration count grows only logarithmically with data size, substantially below the theoretical bound. The practical implementation includes numerical safeguards, and a reference implementation is available as a  Rust package with Python bindings.
The method's low memory footprint also makes it a promising candidate for embedded and real-time systems.

Comprehensive experiments on synthetic and real datasets demonstrate that PALB consistently outperforms existing exact solvers, including Barrodale-Roberts, simplex, and interior-point methods, across all scales. It empirically exhibits log-linear runtime and processes millions of samples in under a second on standard hardware.

Future work will explore efficient updating and downdating strategies for dynamic datasets and extensions to multidimensional LAD regression problems.

  \printbibliography
\end{refsection}

\clearpage
\thispagestyle{empty} 
\onecolumn
\null 
\vfill
\begin{center}
  {\color{gray} Page intentionally left blank}
\end{center}
\clearpage
\twocolumn

\begin{center}
    \Large Supplementary
\end{center}

\begin{refsection}
  In this supplementary we principally give the proofs for the theorems in the main text.

We first show that the subdivision phase of the proposed algorithm terminates with an exact solution of problem~\eqref{eq:main-problem} in finitely many steps (Theorem~\ref{thm:termination}).
Next we give concrete, probabilistic as well as deterministic, bounds on the number of subdivision steps by relating our problem to the $k$-set problem in~\ref{subsec:k-level}. From these we prove the main complexity results.
We prove an upper bound on the number of expansion steps in terms of the initial guess, its distance from some optimal solution as well as the hyperparameters of our method (Proposition~\ref{prop:expansion}). This bound is independent of the number of points.
After stating the subdifferential of the marginal function $J(m) = \min_{t \in \R} f(m,t)$ and giving an algorithm for computing it in section~\ref{subsec:subdiff-J}, we finish with a brief outline of performance profiles and related experimental details in section~\ref{appendix:sec:perfprof}.

\section{Proofs for subdivision steps}

\subsection{Termination with an exact minimizer}

The following lemma is used to show the desired convergence of the subdivision steps.

\begin{lemma}\label{lem:already-minimal-or-new-gradient}
  Let $f : \R \to \R$ be convex.
  Let $a < b$ such that $g_a = \max \partial f(a) < 0 < \min \partial f(b) = g_b$ and let $\xi \in \R$ be the unique point such that

  \begin{align}
    f(a) + g_a (\xi-a) = f(b) + g_b(\xi-b).
  \end{align}

  Then $\xi \in (a,b)$ and exactly one of the following is true:
  \begin{enumerate}
    \item\label{cond:bisect:1} $\xi$ is a global minimizer of $f$.
    \item\label{cond:bisect:2} $g_a < g < 0$ for all $g \in \partial f(\xi)$.
    \item\label{cond:bisect:3} $0 < g < g_b$ for all $g \in \partial f(\xi)$.
  \end{enumerate}
  Moreover in either of the latter two cases it holds that
  $f(a) + g_a (\xi-a) < f(\xi)$;
  in particular $(\xi, f(\xi))$ is not contained in the graph of either of the lines $x \mapsto f(c) + g_c(x - c),\ c =a,b$.
\end{lemma}

Note that the important point here is that $g_a < g$ or $g < g_b$ in cases \ref{cond:bisect:2} and \ref{cond:bisect:3}, and that equality does not hold.
Further, $\xi$ can never be equal to $a$ or $b$ in any case.

\begin{IEEEproof}
  Since $x \mapsto f(t) + g_t (\xi - t), t \in \{a,b\},$ are lines with different slopes they intersect, hence $\xi$ is well defined.
  Let $\lambda = \frac{\xi - a}{b - a}$.
  An elementary calculation shows that for the secant slope $m = \frac{f(b) - f(a)}{b-a}$ we have that
  \begin{align*}
    \lambda &=  \frac{g_b - m}{g_b - g_a},
    \qquad \xi = (1 - \lambda) a + \lambda b.
  \end{align*}
  Observe that since $g_b$ is a subgradient at $b$ we have $m(b-a) + f(a) = f(b) \geq g_a (b-a) + f(a)$ so that $m \geq g_a$.
  Similarly
  $m \leq g_b$.
  Now assume for contradiction that $m = g_a$, then $f$ would be affine linear with slope $m$ on $[a,b]$
  (this follows from the convexity of the function $x \mapsto f(x) - (f(a) + m(x-a))$),
  but then $m$ would also be a subgradient at $b$ and hence $g_b = m = g_a$,
  which contradicts our initial assumption.
  Similarly, we get $m \neq g_b$ so that in total $g_a < m < g_b$.
  Then $g_b - g_b < g_b - m < g_b - g_a$ and hence
  \begin{align*}
    0 < \frac{g_b - m}{g_b - g_a} < 1.
  \end{align*}
  It follows that $\xi$ is a strict convex combination of $a,b$
  and hence is contained in the interior of $[a,b]$.
  
  If $0 \in \partial f(\xi)$ then $\xi$ is a global minimizer of $f$, this is case ~\eqref{cond:bisect:1}.
  Assume that $0 \not\in \partial f(\xi)$.
  Since the subdifferential is always closed convex
  it follows that $\partial f(\xi)$ is either strictly negative or strictly positive.
  Using the monotonicity of the subdifferential and that $a \leq \xi \leq b$ we obtain for any $g \in \partial f(\xi)$, $g_a = \max \partial f(a) \leq  g \leq \min \partial f(b) = g_b$.

  Now assume for contradiction that $g_a \in \partial f(\xi)$,
  then, similarly to before, we find that $f$ is affine with slope $g_a$ on $[a, \xi]$.
  Hence $f(\xi) = f(a) + g_a (\xi - a) \overset{\mathrm{def.}~\xi}{=} f(b) + g_b (\xi - b)$, so that for any $x \in \R$
  \begin{align*}
    f(x) &\geq f(b) + g_b(x - b) = f(\xi) - g_b(\xi - b) + g_b (x - b)
    \\ & = f(\xi) + g_b(x - \xi).
  \end{align*}
  This shows that $g_b \in \partial f(\xi)$. But then $0 \in [g_a, g_b] \subseteq \partial f(\xi)$; a contradiction.
  Similarly one obtains that $g_b \in \partial f(\xi)$ leads to a contradiction.
  The final claim follows directly from the subdifferential inequality,
  using that $0 \not\in \partial f(\xi) \implies g_a \not\in \partial f(\xi)$ as we've just shown.
\end{IEEEproof}

The next lemma intuitively tells us that,
if $f$ is locally given by some line $L$ at $a$ (resp. $b$),
and $\xi$ is not already a minimum of $f$,
then $f$ is necessarily given by some other line in a neighborhood of $\xi$.
So if $\xi$ is not optimal, then the active set $A(\xi) := \{i \in I : f_i(\xi) = f(\xi)\}$ doesn't contain any of the active elements at $a$ or $b$. 

\begin{lemma}\label{lem:already-minimal-or-new-linear-piece}
  Let $f : \R \to \R$ convex and piecewise affine with finitely many pieces ${\{f_i\}}_{i \in I}$ such that $f(x) = \max_{i \in I} f_i(x)$.
  Let $a<b$ and $\xi$ as in Lemma~\ref{lem:already-minimal-or-new-gradient}.
  Then $\xi$ minimizes $f$, or $f(\xi) = \max_{i \in I \setminus A} f_i(\xi)$
  with $A := \{i \in I : f_i(a) = f(a)\} \cup \{i \in I : f_i(b) = f(b)\}$.
\end{lemma}

\begin{IEEEproof}
  The slopes of the lines $\{f_i\}_{i \in A}$ are subgradients of $f$ at $a$ or $b$; and they are active at $\xi$ (i.e.\ $f(\xi) = f_i(\xi)$) if and only if they are subgradients at $\xi$ as well.
  But by Lemma~\ref{lem:already-minimal-or-new-gradient} we know that they cannot be subgradients at $\xi$
  if $\xi$ is not already a minimizer.
\end{IEEEproof}

Note that the \emph{or} in the previous lemma is not an \emph{either or} in general, i.e.\ both can happen.

\begin{thm}\label{thm:termination}
  The subdivision phase of PALB terminates with an exact minimizer of problem~\eqref{eq:main-problem} after a finite number of steps.
  The number of steps is bounded by the number of affine linear segments of the objective function.
\end{thm}
\begin{IEEEproof}
  This follows immediately from the previous lemma~\ref{lem:already-minimal-or-new-linear-piece}: PALB iterates the construction of $\xi$, replacing either $a$ or $b$ by the corresponding $\xi$ along the way.
  In every step at least one element of $I$, i.e.\ a linear segment of the objective function, can be eliminated. Because the minimizer is always contained in the current interval $[a_k ,b_k]$ the claim follows.
\end{IEEEproof}

\subsection{The $k$-level and $k$-set problems}\label{subsec:k-level}

This section and section~\ref{subsec:k-set-for-our-problem} entails the primary work for the proof of the subdivision parts of Theorem~\ref{thm:main-conv} and Theorem~\ref{thm:main-conv-probabilistic}.

To conclude these from lemma~\ref{lem:already-minimal-or-new-linear-piece} it suffices to bound the number of linear segments of $J$.
We do this by relating our problem to a well-studied problem in discrete and combinatorial geometry:
the $k$-set problem, and dually the $k$-level problem \cite{Matousek_2002}.

Given a finite set $P \subseteq \R^2$ of points in the Euclidean plane,
$0 \leq k \leq |P|$,
a \emph{$k$-set for $P$} is a subset $P'$ of size $k$ that can be separated from
the remaining points $P \setminus P'$ by a line.
Dually one is given a set $L \subseteq 2^{\R^2}$ of lines and defines a \emph{$k$-level of the arrangement $\mathcal{A}(L)$}
to be the (for technical reasons) closure of the set of points $p$ on those lines such that exactly $k$ lines are beneath every $p$.
So a $k$-level is essentially a piecewise linear curve formed by segments of the lines $L$,
such that exactly $k$ lines are beneath that curve.
The ``linear pieces'' of this curve are also called \emph{$k$-facets} or \emph{$k$-edges}.
Note that a variety of alternate definitions exists, see for example~\cite{Matousek_2002}.

The $k$-set problem now asks for the maximum number of $k$-sets of a set $P$ of points in general position,
and the $k$-level problem asks for the maximum number of vertices in a $k$-level of a set $L$ of lines in general position;
i.e.\ it seeks to bound the number of kinks in the piecewise linear curve.
These problems are well known to be related via a projective duality:
a point $(x,y)$ is dual to the set of lines passing through that point,
which is in and of itself a line in a parameter space:
it's the set of lines with slope $m$ and intercept $t$ in $\{(m,t) : t = y - mx\}$.
Through such dualities one relates the $k$-set and $k$-level problems and it is well-known
that the solutions to these problems are within a constant factor of oneanother.
Finally, it is known (cf.~\cite{Matousek_2002, Edelsbrunner_1987}) that the situation of the points being in general position is the maximal one:
if the points are not in general position, then there will be no more $k$-sets than if they were.
Overviews for these problems can be found in the seminal paper~\cite{Dey_1997} and the more recent~\cite{Leroux_2022}.

\subsection{Applying the $k$-set problem to our setup}\label{subsec:k-set-for-our-problem}

The connection of the $k$-level / $k$-set problems to the problem we study is as follows:
Denote by $\kLev{N}{k} \in \N_0$ the solution to the $k$-level problem on $N$ points in general position.
It is well known that for any fixed $m \in \R$,
the minimizers of $\min_t \sum_i |mx_i+t - y_i|$ are precisely given by the set of medians $\Median {(y_i - m x_i)}_{i}$.
Under the duality from above this means we have $N$ lines given as the graphs of $p^*_i(m) := y_i - m x_i$,
and we are interested in the closed region $\MedReg := \{(m,t) : t \in \Median {(y_i - m x_i)}_{i}\}$.
Because of the convexity of the original problem the set of minimizers is convex as well, and hence $\MedReg \cap \{m = m_0\}$ is convex for any $m_0$.

As $m$ varies across $\R$, the corresponding medians can only change at points where two lines $p^*_i, p^*_j$ intersect (and thus swap positions in the order),
and between two such intersections the objective function is necessarily linear.
Hence, the points we are interested in for our complexity analysis are intersections of the lines ${\{p^*_i\}}_i$ that are in $\MedReg$.

We consider the \emph{lower and upper median curves} $\LoMed$, $\HiMed$ defined as the piecewise linear curves consisting of
all $(m,t)$ such that $t$ is the minimal, resp.\ maximal, value $t'$ with $(m,t') \in \MedReg$.
Note that whenever $N$ is odd, both of these curves coincide since all the medians defining $\MedReg$ are unique in this case;
in this case we refer to $\LoMed = \HiMed$ as the median curve.

Let us begin by discussing this simpler case:
Outside of the points of intersection the medians are unique,
and hence there must be exactly $(N - 1) / 2$ lines (out of the ${\{p^*_i\}}_{i=1,...,N}$) below the
median curve at every $m$ where no two lines intersect.
The points of intersection themselves are limit points of $\MedReg$.
And finally whenever there are exactly $(N - 1) / 2$ lines below some point $(m,t)$ on any of the lines,
that point must be the unique median or a point of intersection and hence is contained in $\MedReg$.
This shows that $\MedReg$ is precisely the $\left((N-1) / 2\right)$-level of the arrangement associated to the lines ${\{p_i^*\}}_i$.
Hence the number of vertices of $\MedReg$ and thus the number of ``kinks'' of $J$ is given by the solutions to the $k$-level problem $\kLev{N}{(N-1)/2}$.

If on the other hand $N$ is even, then there are $N / 2 - 1$ lines below the low median curve and $N/2$ below the high median curve outside of the points of intersections.
Arguing similarly to above we find that indeed the low median curve is the $\left(N/2 - 1\right)$-level while the high median is the $(N/2)$-level.
We can hence give a (quite crude, compared to the odd case) bound on the number of vertices of $\MedReg$ by the sum $\kLev{N}{N/2 - 1} + \kLev{N}{N/2}$.

To handle both cases of even and odd $N$ uniformly we bound
\begin{align}
  \# \operatorname{Vertices}(\MedReg) \leq 2 \kLevBound{N}{N/2} \label{eq:vert-rough-bound}
\end{align}
in our following analysis; where we are using that the known upper bounds $\kLevBound{N}{\cdot}$ for the $k$-level problem are monotonuous in $k$.

\begin{prop}\label{prop:subdivision-part-main-conv}
  The $O(N^{4/3})$ bound on the number of subdivision steps of Theorem~\ref{thm:main-conv} holds.
\end{prop}
\begin{IEEEproof}
  By the preceeding explanation this follows from the bound $O(Nk^{1/3})$ for the $k$-level problem obtained in~\cite{Dey_1997}:
  by Lemma~\ref{lem:already-minimal-or-new-linear-piece} it suffices to bound the number of linear segments of $J$.
  This number is bounded by the number of vertices of the median region $\MedReg$, which in turn can be bounded by $2 \kLevBound{N}{N/2}$, where \cite{Dey_1997} gives $\kLevBound{N}{N/2} \in O(N(N / 2 + 1)^{1 / 3}) = O(N^{4 / 3})$.
\end{IEEEproof}

\begin{prop}\label{prop:subdivision-part-main-conv-probabilistic}
  The $\frac{C N^{7/4}}{{(N/2 + 1)}^{1/4} {(N/2 - 2)}^{1/4}} \in O(N^{5/4})$ bound on the number of subdivision steps of Theorem~\ref{thm:main-conv} holds.
\end{prop}
\begin{IEEEproof}
  This works analogously to the deterministic case.
  We dualize the $k$-level problem to a $k$-set problem and apply the results of~\cite[Theorem 3.2]{Leroux_2022}:
  we map each point $(x,y)$ into the line $t = y - mx$ to state the $k$-level problem we are interested in,
  and then back into a point $(-x,-y)$ via the standard duality $\mathcal{D} : \{(m,t) : t = am - b\} \mapsto (a,b)$ of~\cite[Definition 5.1.4]{Matousek_2002}.

  To apply the $k$-set result of~\cite{Leroux_2022} we have to discuss the distribution of the point samples after the dualization:
  The points $(x,y)$ follow a distribution which admits a density precisely when the image points $(-x,-y)$ of the $k$-set problem do:
  let the original measure $\mu$ have density $u$,
  then the dualized points follow the pushforward measure $\nu(A) := \int_{A} u(-x) \diff x$ with density $u \circ (- \cdot)$.

  Now~\cite{Leroux_2022} gives a bound $\kLevBound{N}{k} = \frac{58 N^{7/4}}{{(k + 1)}^{1/4} {(N - 2 - k)}^{1/4}}$.
  Inequality~\eqref{eq:vert-rough-bound} incurs an extra factor of 2 on top of that, and there is an additional(generally unknown) constant factor that comes from relating the $k$-level and $k$-set problems. Together these constitute the constant $C$ in the claimed bound.
\end{IEEEproof}

Finally let us remark that it was conjectured by Erdős that the number of $k$-edges of $N$ points in
the plane is $O(N^{1+\varepsilon})$ for any $\varepsilon > 0$,
and that there has since been some evidence suggesting that this may indeed be true according to~\cite{Leroux_2022}.
Future work on this still open problem may hence allow giving better bounds for the complexity of our method.

\section{Proofs for expansion steps}

The expansion step part of Theorem~\ref{thm:main-conv} follows from the following Proposition:

\begin{prop}\label{prop:expansion}
  Let $M^* = [m^*_-,~m^*_+]$ be the solution set to $\min_m J(m)$.
  Denote by $d_{M^*}(x) = \inf_{m \in M^*} |x - m|$ the distance function of $M^*$.
  Let $m_0 \in \R, \mu \in (0,~\infty)$.
  Then our method takes no more than $K$ expansion steps with
  \begin{align}
    K \leq \log_2 \left( \frac{d_{M^*}(m_0)}{\mu |m_0|} + 1 \right) - 1
  \end{align}
  and the length $\Delta_K$ of $M_K = [a_K, b_K]$ is bounded by
  \begin{align}
    \Delta_K \leq \frac{1}{2} \left( d_{M^*}(m_0) + \mu |m_0| \right)
  \end{align}
\end{prop}
\begin{IEEEproof}
  If $m_0 \in M^*$ the algorithm terminates, hence assume that $m_0 \not\in M^*$.
  If $m^*_+ < a_0$ the algorithm starts with ``leftward'' expansion for $K$ steps.
  Unwinding recursively we find that during this expansion
  \begin{align*}
    a_{k+1} &= a_k - \delta_k = a_{k-1} - \delta_{k-1} - \delta_k
    \\ ...  &= a_0 - \sum_{r=0}^k \delta_k = m_0 - \mu |m_0| - \sum_{r=0}^k 2^{r+1} \mu |m_0|
    \\  &= m_0 - \mu |m_0| (2^{k+2} - 1).
  \end{align*}
  This is strictly falling in $k$, so it's clear that at some point $a_{k+1} \leq m^*_+$;
  in fact this happens once
  \begin{align*}
    a_{k+1} = m_0 - \mu |m_0| (2^{k+2} - 1) \leq m^*_+
    \\ \iff \log_2 \left( \frac{m_0 - m^*_+}{\mu |m_0|} + 1 \right) - 2 \leq k,
  \end{align*}
  so after $K_- := \ceil{\log_2 \left( \frac{m_0 - m^*_+}{\mu |m_0|} + 1 \right) - 2}$ expansion steps.
  In the same manner we obtain that when $b_0 < m^*_-$ the algorithm terminates after
  $K_+ := \ceil{\log_2 \left( \frac{m^*_- - m_0}{\mu |m_0|} + 1 \right) - 2}$ expansion steps.
  This yields the desired upper bound for $K$
  \begin{align*}
    K &\leq \log_2 \left( \frac{d_{M^*}(m_0)}{\mu |m_0|} + 1 \right) - 1
    \\ \text{where}~d_{M^*}(m_0) &= \min \{ m_0 - m^*_+, m^*_- - m_0 \}
  \end{align*}
  so that the length $\Delta_K = 2^K \mu |m_0|$ of the segment $[a_K, b_K]$ is bounded by
  \begin{align*}
    \Delta_K \leq \frac{1}{2} \left( d_{M^*}(m_0) + \mu |m_0| \right).
  \end{align*}
\end{IEEEproof}

\begin{IEEEproof}[Proof of Theorem~\ref{thm:main-conv}]
  Termination with an exact minimizer in finitely many steps follows from Theorem~\ref{thm:termination}. The bound on the number of subdivision steps is Proposition~\ref{prop:subdivision-part-main-conv} as described in section~\ref{subsec:k-set-for-our-problem}.
  The bound on the number of expansion steps follows from~\ref{prop:expansion}.
\end{IEEEproof}

\begin{IEEEproof}[Proof of Theorem~\ref{thm:main-conv-probabilistic}]
  The termination and exactness claims are the same as in Theorem~\ref{thm:main-conv}.
  The claimed probabilistic bound on the number of subdivision steps is Proposition~\ref{prop:subdivision-part-main-conv-probabilistic} as described in section~\ref{subsec:k-set-for-our-problem}.
\end{IEEEproof}

\section{Computing the subdifferential of the objective function}\label{subsec:subdiff-J}

Let $(x_i,y_i)_{i=1,...,N}$ be points in the plane
and
\begin{align}
  J(m) = \min_t \sum_{i=1}^N |y_i - (mx_i + t)| \label{sup:eq:recall-J}
\end{align}
as in the main text.
For any $m,t \in \R$ define the index sets
\begin{align*}
  I_+ &= \{i : x_i m + t - y_i > 0\},
\\ I_- &= \{i : x_i m + t - y_i < 0\},
\\ I_0 &= \{i : x_i m + t - y_i = 0\}.
\end{align*}
For brevity the dependence on $m,t$ is not made explicit at this point.
Further denote by $\bar{t} = \max \Median {(y_i - m x_i)}_i$ and $\underbar{t} = \min \Median {(y_i - m x_i)}_i$ the upper and lower medians at some $m \in \R$ respectively.

The following Lemma is a standard result \cite{yamamoto1988algorithms, megiddo1984linear} that we repeat here for convenience.

\begin{lemma}
    The function $J$ given in~\eqref{sup:eq:recall-J} is a convex piecewise linear function.
    Denote by $M$ the set of kinks of $J$.
    Then $J$ is differentiably outside of $M$,
    and its derivative at $m \not\in M$ is given by
		\begin{align}
			g'(m) =   \sum_{i \in I_+(\bar{t})} x_i - \sum_{i \in I_-(\underbar{t})} x_i.	
		\end{align}
\end{lemma}

One issue that arises when applying this lemma in practice is that it's not necessarily
perfectly clear how to use it to compute the subdifferential at a kink. Here one has to consider the values $g'(m - \varepsilon), g'(m + \varepsilon)$ for small enough $\varepsilon > 0$ to, in effect, compute the two directional derivatives $J'(m; -1), J'(m; 1)$ that give the bounds for the subdifferential.

We now state a proposition that gives a generally applicable expression for the subdifferential that remains valid inside the kinks of $J$. We then describe a linear-time method for computing this expression.

\begin{prop}
  The subdifferential of $J$ is given by
  \begin{align*}
    \partial J(m) = E_m(\bar{t}) \cup E_m(\underbar{t}), \quad m \in \R
  \end{align*}
  where the sets $E_m(t)$ are defined by
  \begin{align*}
    E_m(t) &= \{S + \sum_{i \in I_0} \alpha_i x_i : \Norm{{(\alpha_i)}_{i \in I_0}}_\infty \leq 1, \sum_{i \in I_0} \alpha_i = -B \},
    \\ S &= \sum_{i \in I_+} x_i - \sum_{i \in I_-} x_i, \qquad B = |I_+| - |I_-|.
  \end{align*}
  The sets $E_m(t)$ are compact intervals whose boundaries $s_{\min} + S, s_{\max} + S$ are given by
  \begin{align*}
    s_{\min} &= \min_{(\alpha_i)_{i \in I_0} \in {[-1,1]}^{|I_0|}} \sum_{i \in I_0} \alpha_i x_i \qquad \text{s.t.}~ \sum_{i \in I_0} \alpha_i = -B
    \\ s_{\max} &= \max_{(\alpha_i)_{i \in I_0} \in {[-1,1]}^{|I_0|}} \sum_{i \in I_0} \alpha_i x_i \qquad \text{s.t.}~ \sum_{i \in I_0} \alpha_i = -B.
  \end{align*}
\end{prop}

We omit the proof of the proposition at this point since it is a lengthy but standard calculation using standard results of convex and variational analysis \cite[Theorem 4.2.1, Chapter VI]{HiriartUrruty1993i}, \cite[Theorem 10.13]{Rockafellar1998}.

To compute $\partial J(m)$ one first determines the index sets $I_-, I_+, I_0$ via introselects. From this we compute the sum $S$. Computing the two additional values $s_{\min}, s_{\max}$ then boils down to a simple selection problem (that can also be solved with an introselect) as we describe next.

It's convenient to begin by transforming the linear programs that define $s_{\min}$ and $s_{\max}$:
the change of variables $\beta_i = \frac{\alpha_i + 1}{2} \in [0,1]$ in the LP defining $s_{\max}$ yields
\begin{align*}
  &\max_{(\beta_i)_{i \in I_0} \in {[0,1]}^{|I_0|}} \sum_{i \in I_0} (2 \beta_i - 1) x_i \quad \text{s.t.}~ \sum_{i \in I_0} (2 \beta_i - 1) = -B \\
  \Leftrightarrow  
  &\max_{(\beta_i)_{i \in I_0} \in {[0,1]}^{|I_0|}} \sum_{i \in I_0} \beta_i x_i \quad \text{s.t.}~ \sum_{i \in I_0} \beta_i = \frac{|I_0| - B}{2}. \tag{K} \label{eq:knacksack-specific}
\end{align*}
This is an instance of the continuous, equality knapsack problem \cite{Fravel2024}.
The optimal objective value $O_\beta$ of this problem can be transformed back to the one of the original problem via $O_\alpha = 2 O_\beta - \sum_{i \in I_0} x_i$.
Similarly $s_{\min}$ may be found with the analogous minimization problem.

Whenever $I_0$ is a singleton the solution to problem \eqref{eq:knacksack-specific} is trivial.
The general problem can be solved in linear time using a greedy selection algorithm:
consider the model problem
\def\kpvar{v}
\begin{align*}
  \max_{(\beta_i)_{i \in \mathcal{I}} \in {[0,1]}^{|\mathcal{I}|}} \sum_{i \in \mathcal{I}} \beta_i \kpvar_i \qquad \text{s.t.}~ \sum_{i \in \mathcal{I}} \beta_i = C
\end{align*}
for some values ${(\kpvar_i)}_{i \in \mathcal{I}}$ and \emph{capacity} $C \geq 0$.
Assume for simplicity that $\mathcal{I} = 1,...,\mathcal{N}$
and begin by considering the case when $\kpvar_1 \geq \kpvar_2 \geq ... \geq \kpvar_{\mathcal{N}}$.
Then set $\beta_{i} = 1$ while $C - \sum_{j=1}^{i-1} \beta_j = C - (i-1) \geq 1$, and assuming this process stops at index $\hat{i}$,
set $\beta_{\hat{i}} = C - \sum_{j=1}^{\hat{i}-1} \beta_j = C - (\hat{i}-1)$ and $\beta_{i} = 0$ for all remaining $i$.
Hence the problem is reduced to that of determining the \emph{pivot} index $\hat{i}$
--- i.e. the minimal index where the condition $C - (\hat{i}-1) \geq 1$ first fails.
We then have $\hat{i} = \min \{i : C - (i - 1) < 1 \} = \min \{i : C < i \}$,
so that the pivot index is $\hat{i} = \floor{C} + 1$.

To solve the knapsack problem for potentially unsorted $\kpvar_i$,
one must first determine the pivot index $\hat{i} = \floor{C} + 1$.
The goal is then to find the element that would be at this index in the sorted collection,
which is possible in linear time using a selection algorithm like introselect.
This algorithm simultaneously partitions the collection into elements with values $z_i$ larger than the pivot and those with smaller values.
One can then calculate the weighted sum where weights $\beta_i$ are set to $1$ for the \emph{larger} group,
$0$ for the \emph{smaller} group,
and a fractional value in $[0,1]$ for the pivot element itself (this ``fractional'' value won't ever be actually fractional in our specific case, since our budget $B$ is an integer).

\section{Performance Profiles}\label{appendix:sec:perfprof}

Let $\mathcal{S}$ be a set of solvers and $\mathcal{P}$ a set of problems (both finite).
The performance profile associated to those two sets and an associated collection of \emph{scores} is defined as follows:

The score (in our case this is the runtime of the method on the one hand, and the final value of the objective function on the other hand) of solver $s \in \mathcal{S}$ on problem $p \in P$ is denoted $t_{p, s}$ and it is assumed that a lower score is ``better''.
For each problem $p$ and any solver $s$ we assign a value $r_{p,s}$ based on how much worse the method performed relative to the best one on this particular problem:
\begin{align*}
  r_{p,s} = \frac{t_{p,s}}{\min_{s' \in \mathcal{S}} t_{p, s'}};
\end{align*}
if some method does not solve a problem then we set $r_{p,s} = \infty$.

To evaluate the overall performance of solver $s$ on the full set $\mathcal{P}$ we evaluate the function
\begin{align*}
  \rho_s : \R_{\geq 1} \to [0,1], \tau \mapsto \frac{|\{p \in \mathcal{P} : r_{p, s} \leq \tau\}|}{|\mathcal{P}|}.
\end{align*}
This function is the performance profile associated to $(\mathcal{S}, \mathcal{P}, t_{\cdot, \cdot})$.
Intuitively the value $\rho_s(\tau)$ gives the probability that the score of solver $s$ on a randomly chosen problem $p$ from $\mathcal{P}$ is at least within a factor $\tau$ of the best value among all the solvers from $\mathcal{S}$.
So, in particular, the value $\rho_s(1)$ gives the probability that $s$ attains the best possible value among all solvers in $\mathcal{S}$.
Generally it is desirable for a method $s$ to have a high value (equal or close to $1$) of $\rho_s(\tau)$ for small $\tau$, since this indicates that it solves many problems from the considered set well.

We consider the four performance profiles $(\mathcal{S}_\text{time}, \mathcal{P}, \text{runtime})$ and $(\mathcal{S}_\text{obj}, \mathcal{P}, \text{final objective value})$
with $\mathcal{S}_\text{time}$ consisting of all tested methods, and $\mathcal{S}_\text{obj}$ consisting of all except for CPLEX, and $\mathcal{P} \in \{\mathcal{P}_\text{synth}, \mathcal{P}_\text{real}\}$.
The reason for CPLEX being excluded from the objective value results is that the CPLEX API we used did not expose primal solutions to the LP.

The problem set $\mathcal{P}_\text{synth}$ used for the synthetic-data performance profile consists of all synthetic series with at most $100\,000$ samples.
We chose to do this because we ran the methods only up to the point where they (approximately) started exceeding 1 second of runtime for the first time --- and the lowest such point was at approximately $N=100\,000$.
This cut-off was necessary to avoid prohibitively long benchmark times because of our large test dataset,
however it complicates the current analysis: it means that past the cut-off point a method is considered to not have solved any problems,
so if we didn't cut off the performance profile dataset at this point, it would result in an inaccurate representation of the given method.

We should say that this relatively early cut-off benefits methods that do very well on ``small'' data,
while being disadvantageous to methods that only start doing well on really big data.
In particular we will see that the IRLS method implemented by statsmodels performs quite poorly according to the performance profile when compared to the various other methods;
however this hides that fact that it still does well on very large datasets.
We elected to still go this route because even on the larger data, where only our method and IRLS were able to complete in-time, our method still outperformed IRLS and as such we don't think that the cut-off at $100\,000$ gives a wrong impression as far as the purposes of our paper are concerned.

The experiment on real data can be considered more generally representative, since in this case all methods run on the full dataset.
  \printbibliography
\end{refsection}

\end{document}